\documentclass[lettersize,journal]{IEEEtran}
\usepackage{amsmath,amsfonts}
\usepackage{algorithm}
\usepackage{algpseudocode}
\usepackage{hyperref}
\usepackage{array}
\usepackage{courier}
\usepackage{xcolor}
\usepackage{subfigure}
\usepackage{textcomp,soul}
\usepackage{stfloats}
\usepackage{url}
\usepackage{verbatim}
\usepackage{graphicx}
\usepackage{booktabs}
\usepackage{multirow}
\usepackage{cite}
\usepackage{amsthm}
\usepackage{siunitx}
\usepackage{bm}
\usepackage{pifont}
\usepackage{threeparttable}
\usepackage{tabularx}
\usepackage{cuted}
\usepackage{algorithm}
\usepackage{algpseudocode}
\usepackage{xcolor}
\usepackage{amsmath,amssymb}
\providecommand{\Return}{\textbf{return}\ }
\usepackage[font=footnotesize]{caption}
\definecolor{lblue}{rgb}{0, 0, 0}

\theoremstyle{definition}

\newcommand{\MYhref}[3][blue]{\href{#2}{\color{#1}{#3}}}%

\makeatletter
\pretocmd\@bibitem{\color{black}\csname keycolor#1\endcsname}{}{\fail}
\newcommand\citecolor[1]{\@namedef{keycolor#1}{\color{blue}}}
\makeatother 
\title{\textcolor{lblue}{UniManip: Open-World Robotic Manipulation via Agentic Operational Graphs}}
\author{Haichao Liu, Yuanjiang Xue, Yuheng Zhou, Li Yang, Haoyuan Deng, Yinan Liang, Lihua Xie, and Ziwei Wang
\thanks{Haichao Liu, Yuanjiang Xue, Yuheng Zhou, Li Yang, Haoyuan Deng, Lihua Xie, and Ziwei Wang are with the School of Electrical and Electronic Engineering, Nanyang Technological University, Singapore (e-mail: haichao.liu@ntu.edu.sg; e250086@e.ntu.edu.sg; yuheng.zhou@ntu.edu.sg; lyang018@e.ntu.edu.sg; haoyuan.deng@ntu.edu.sg; elhxie@ntu.edu.sg; ziwei.wang@ntu.edu.sg).}
\thanks{Yinan Liang is with the Department of Automation, Tsinghua University, China (email: liangyn24@mails.tsinghua.edu.cn)}
\thanks{This work has been submitted to the IEEE for possible publication. Copyright may be transferred without notice, after which this version may no longer be accessible.}
}
\begin{document}
\maketitle
\begin{strip}
\vspace{-3.7cm} %
    \centering
    \includegraphics[width=1\linewidth]{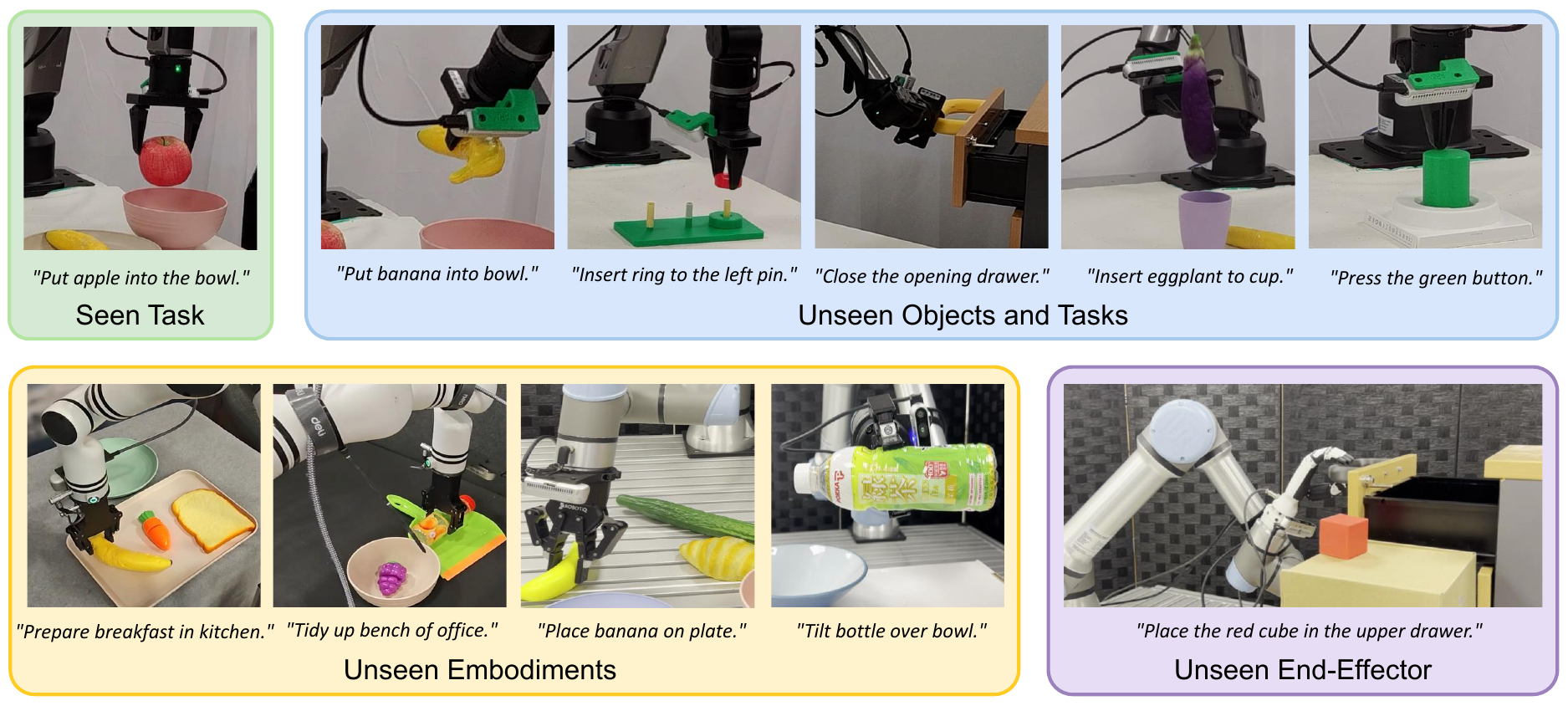}
    \captionof{figure}{\textcolor{lblue}{UniManip enables zero-shot open-world manipulation by grounding language-conditioned tasks into a physically grounded operational graph with closed-loop failure verification. The framework generalizes across unseen tasks, objects, and diverse robotic embodiments, including fixed-base arms, mobile manipulators, parallel-jaw grippers, and dexterous hands.}}
    \label{fig:teaser}
\end{strip}
\begin{abstract}
Achieving general-purpose robotic manipulation requires robots to seamlessly bridge high-level semantic intent with low-level physical interaction in unstructured environments. However, existing approaches falter in zero-shot generalization: end-to-end Vision-Language-Action (VLA) models often lack the precision required for long-horizon tasks, while traditional hierarchical planners suffer from semantic rigidity when facing open-world variations. 
\textcolor{lblue}{To address this, we present UniManip, a unified framework centered on an Agentic Operational Graph (AOG) that serves as an explicit cognitive and operational backbone for task orchestration, state tracking, execution verification, and closed-loop failure recovery. 
Here, high-level semantic intents are grounded into actionable primitives via a physically grounded scene graph, which provides geometric affordances, contact-aware interaction parameters, and collision-free trajectory synthesis.} By coupling a high-level Agentic Layer for task orchestration with a low-level Scene Layer for dynamic state representation, the system continuously aligns abstract planning with geometric constraints, enabling robust execution. Unlike static pipelines, UniManip operates as a dynamic agentic loop: it actively instantiates object-centric scene graphs from unstructured perception, parameterizes these representations into collision-free trajectories via a safety-aware local planner, and exploits structured memory to autonomously diagnose and recover from execution failures. \textcolor{lblue}{UniManip achieves state-of-the-art performance on the LIBERO-Pro benchmark, while extensive real-world experiments demonstrate robust zero-shot generalization across diverse tasks, embodiments, and dexterous hands, outperforming leading VLA models and hierarchical baselines by 22.5\% and 25.0\% in task success rate, respectively. Our open-source project page can be found at \MYhref{https://henryhcliu.github.io/unimanip}{https://henryhcliu.github.io/unimanip}.}
\end{abstract}


\section{Introduction}
\label{sec:introduction}
General-purpose robotic manipulation aims to deploy agents into unstructured, open-world environments where they must execute diverse tasks without pre-configuration. Central to this utility is zero-shot generalization, the capacity to instantly adapt to novel objects and layouts without task-specific training or fine-tuning~\cite{billard2019trends, zhang2025zisvfm}. However, replicating human-level adaptability remains a formidable challenge due to the fundamental disconnect between learned priors and novel realities. The primary bottleneck in current systems is the out-of-distribution (OOD) condition: end-to-end models often fail when facing unseen visual data or free-form human commands that deviate from training distributions, whereas traditional hierarchical planners lack the semantic flexibility to handle unexpected faults. Consequently, achieving robust zero-shot performance demands more than open-loop execution; it requires a fundamental reasoning ability, specifically, an agentic capacity to continuously perceive, verify, and autonomously reflect to realign high-level intent with the unscripted physical world~\cite{abou2025agentic, guo2025embodied}.

To approach zero-shot generalization, current research predominantly follows two paradigms: data-driven Vision-Language-Action (VLA) adaptation~\cite{zitkovich2023rt, kim2025openvla, black2024pi_0, hung2025nora, amin2025pi} and structured hierarchical planning~\cite{huang2025rekep, huang2023voxposer, liu2025robodexvlm, alayrac2022flamingo, chenfast,jiao2025integration}. In the VLA domain, strategies like in-context learning~\cite{zhao2025mos,kamil2025learning} and parameter-efficient fine-tuning~\cite{li2025controlvla} attempt to steer frozen priors toward new tasks. However, these methods remain fundamentally bounded by their backbone's training distribution, struggling to generalize to unseen kinematics or out-of-distribution dynamics without high-quality demonstrations. Conversely, hierarchical frameworks~\cite{huang2025rekep, huang2023voxposer, liu2025robodexvlm, alayrac2022flamingo, chenfast,jiao2025integration} leverage VLMs or PDDL to decompose tasks into symbolic plans, yet they typically treat planning and execution as isolated stages. This separation results in brittleness, as high-level logic lacks the dynamic physical grounding and closed-loop feedback required to adapt when primitives fail~\cite{huang2025rekep}. This dilemma leads to a critical research question: \textit{How can we efficiently represent scenario conditions and spatial operations to serve VLM reasoning while seamlessly bridging it with precise, low-level manipulation execution?}

\textcolor{lblue}{To bridge this divide, we propose UniManip, a physically grounded framework built around a Bi-level Agentic Operational Graph (AOG). The upper Agent Logic Graph (ALG) is the procedural component: it represents operation sequencing, tool selection, verification, and recovery conditions. The lower Semantic-Operational State Graph (SOSG) is the episodic component: it represents task-relevant 3D object/part state, spatial relations, and manipulation affordances. Coupling these layers makes the operation context persistent across perception, planning, execution, and recovery: verification updates the SOSG, and the ALG uses the updated state to repair or replan the remaining sequence.}
\begin{figure*}[t]
    \centering
    \includegraphics[width=0.95\textwidth]{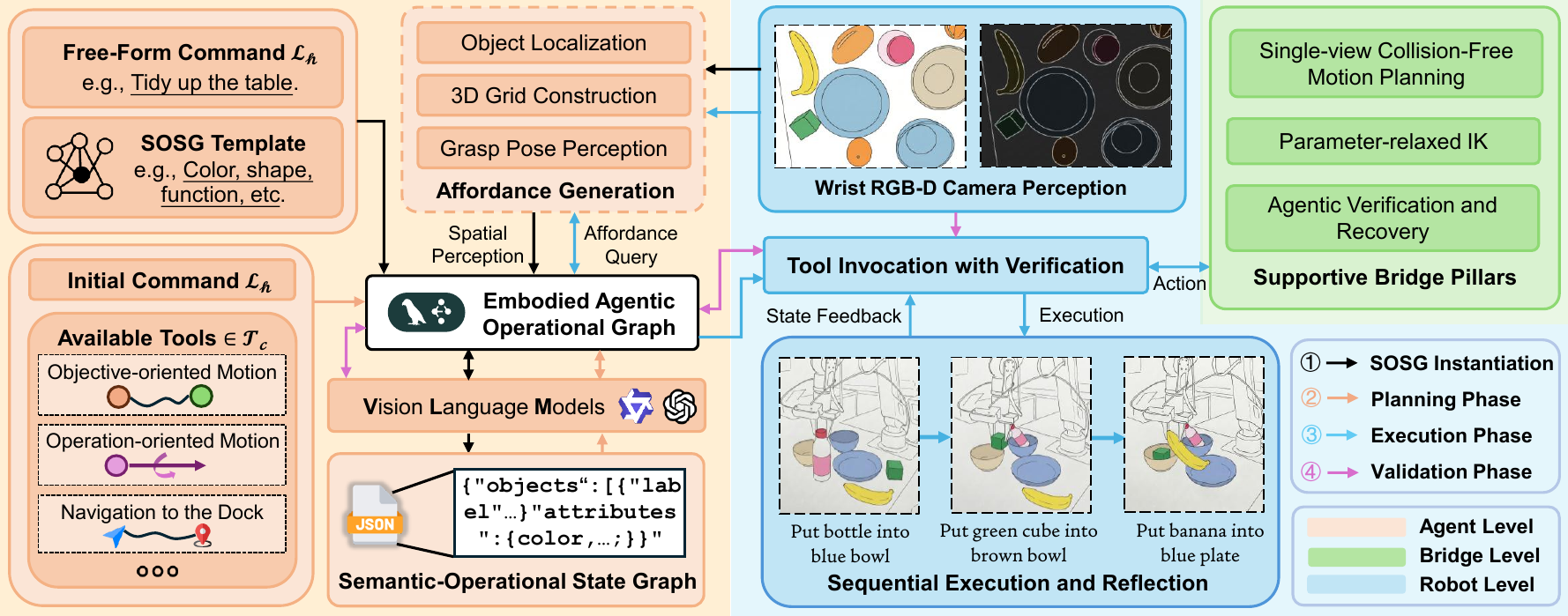}
    \caption{Overview of the UniManip framework. The system integrates high-level task planning with low-level motion execution through an Agentic Operational Graph (AOG), illustrated at the agent level. The VLM interprets human commands to generate an operational graph, which guides the robot's actions. A reflective recovery mechanism allows the system to diagnose and adapt to execution failures.}
    
    \label{fig:overview}
\end{figure*}

As illustrated in Fig.~\ref{fig:overview}, this agentic framework operates through three logically connected phases: First, multimodal scene instantiation, which instantiates multimodal inputs into object-centric geometric attributes and physical affordances. Second, agentic primitive execution, which generates skill primitives and drives action parameterization via a single-view conservative reconstruction approach, utilizing safety-aware objectives to ensure collision-free trajectories. Third, closed-loop reflection and recovery, which exploits the AOG’s structured memory to autonomously diagnose execution anomalies and dynamically restructure the logic graph for recovery.

The main contributions are summarized as follows:

\begin{itemize}
    \item \textcolor{lblue}{We propose UniManip, an agentic operational-graph framework that grounds semantic task intent into 3D scene state, interaction-relevant affordances, and collision-free motion for execution without task-specific fine-tuning in the evaluated settings.}
    
    \item \textcolor{lblue}{We introduce the Bi-level AOG, an explicit execution representation combining procedural memory for operation-level control flow with episodic memory for time-indexed states, verifications, and recovery context.}
    
    \item We establish a robust task-to-motion bridge that combines single-view conservative reconstruction with a safety-aware planner, enabling the generation of optimized, collision-free, and efficient end-effector trajectories.
    
    \item \textcolor{lblue}{We evaluate against VLA, hierarchical, graph-based, and matched no-graph baselines, and validate transfer to mobile, parallel-jaw, and multi-fingered embodiments with the corresponding embodiment adapters.}
\end{itemize}

\section{Related Work}
\label{sec:related work}
\textcolor{lblue}{We review related work in the following areas: end-to-end methods without reasoning, hierarchical, graph-based, and code-generation methods that plan before acting, and failure detection and recovery in robotic manipulation.}
\subsection{VLA Models for Manipulation}
This category of methods maps raw sensory observations directly to low-level robot actions, leveraging the scaling laws of large imitation learning datasets~\cite{o2024open}. Representative examples include RT-1~\cite{brohan2023rt}, PaLM-E~\cite{driess2023palm}, OpenVLA~\cite{kim2025openvla}, $\pi_0$~\cite{black2024pi_0}, and GR00T~\cite{bjorck2025gr00t}. These models typically fuse visual and textual inputs within a unified transformer architecture to generate motor commands.
Beyond fully offline training, recent work has explored few-shot adaptation of pretrained VLAs to new tasks. MoS-VLA~\cite{zhao2025mos} demonstrates one-shot skill adaptation, while \textit{Learning a thousand tasks in a day}~\cite{kamil2025learning} scales rapid task acquisition through large-scale multi-task learning and transfer. ControlVLA~\cite{li2025controlvla} further improves few-shot generalization via object-centric adaptation of pretrained VLA backbones. Moreover, Nora 1.5~\cite{hung2025nora} leverages reward-guided post-training to achieve high-fidelity zero-shot performance across unseen tasks and object categories.

Despite strong performance in seen or lightly adapted settings, these methods often behave as ``black boxes" with limited explicit state grounding and minimal built-in mechanisms for long-horizon verification and recovery~\cite{deng2025survey}, making them sensitive to distribution shifts and difficult to interpret.

\subsection{Hierarchical Methods with Planning before Action}
To compensate for the limited explicit reasoning in end-to-end VLAs, recent work decouples high-level planning from low-level execution. These approaches can be broadly divided into sub-task guided VLA models~\cite{black2025pi_,zhao2025cot,zawalski2025robotic,team2025gigabrain} and hierarchical reasoning-control frameworks~\cite{li2025hamster,huang2023voxposer,liu2025robodexvlm,alayrac2022flamingo,liu2024moka,huang2024copa,huang2025rekep,pan2025omnimanip}.

Sub-task guided VLA models introduce an explicit planning or reasoning stage before producing actions (often via an action tokenizer). For instance, $\pi_{0.5}$~\cite{black2025pi_} uses a pretrained VLM to generate sub-tasks that are subsequently executed by a fine-tuned VLA. CoT-VLA~\cite{zhao2025cot} incorporates a visual chain-of-thought mechanism by autoregressively predicting future frames as visual goals that guide short-horizon action generation. Similarly, ECoT~\cite{zawalski2025robotic} trains VLAs to perform multi-step reasoning over plans, sub-tasks, and grounded intermediates (e.g., object bounding boxes) before predicting robot actions. GigaBrain-0~\cite{team2025gigabrain} emphasizes language-based subgoal generation and jointly optimizes trajectory regression, subgoals, and discrete action tokens. Despite improved interpretability at the planning stage, these methods are often data-hungry and can remain brittle under OOD conditions.

Methods that bridge high-level reasoning with low-level geometric control typically introduce structured intermediate representations (e.g., value maps, keypoints, or constraints) to ground language into executable motions. Hamster~\cite{li2025hamster} fine-tunes a high-level VLM to produce a coarse 2D path that guides a low-level, 3D-aware control policy. VoxPoser~\cite{huang2023voxposer} leverages the code-writing capabilities of Large Language Models (LLMs) to compose 3D value maps grounded in the observation space, enabling model-based planning of closed-loop trajectories. \textcolor{lblue}{CaP-X~\cite{fu2026cap} similarly studies LLM code-generation agents for manipulation through standardized task APIs and perturbations.} MOKA~\cite{liu2024moka} uses mark-based visual prompting to represent affordances as point sets, converting keypoint prediction into VLM-friendly queries. CoPa~\cite{huang2024copa} decomposes manipulation into task-oriented grasping and task-aware motion planning by extracting part-level geometric constraints for post-grasp poses. ReKep~\cite{huang2025rekep} represents tasks as relational keypoint constraints expressed as functions and solves them via hierarchical optimization. Finally, OmniManip~\cite{pan2025omnimanip} proposes a zero-training approach based on object-centric canonical spaces and functional affordances.
\textcolor{lblue}{Graph-based approaches provide another structured interface between perception and action. RoboEXP~\cite{roboexp2024}, for example, uses an action-conditioned scene graph to support interactive exploration and manipulation by associating object states with actions. Such representations complement geometric planners by retaining state and relational context across interactions.}

While promising, these approaches face practical challenges in open-world deployment. Many rely on multi-view sensing or careful scene setups that are difficult to maintain, especially for mobile manipulators. In addition, keypoint- and constraint-based formulations often require embodiment-specific optimization and tuning, which can hinder cross-embodiment transfer and long-horizon robustness. Finally, failure recovery is often limited because verification and causal analysis are not deeply integrated into the planning-execution loop.

\subsection{Manipulation Failure Detection and Autonomous Recovery}
Robustness to execution failures is essential for long-horizon robotic manipulation~\cite{li2025gr, wu2020squirl}, particularly under zero-shot conditions. Existing work on failure detection and recovery can be broadly categorized into data-driven and model-based approaches. Data-driven methods learn to detect or recover from failures using sensory feedback, e.g., via reinforcement learning for adaptive behaviors or supervised classification of failure modes. A modular hierarchical framework is proposed by decoupling manipulation into base task-oriented skills and reinforcement learning (RL)-based failure prevention skills~\cite{ak2023learning}. Moreover, RecoveryChaining~\cite{vats2025recoverychaining} employs hierarchical RL to learn recovery policies that, when triggered by failure detection, return the robot to a state where nominal model-based controllers can resume the task. In a supervised setting, FINO-Net~\cite{inceoglu2023multimodal} utilizes multimodal sensor fusion to detect anomalies, though it relies on datasets restricted to a limited set of manually classified and rigid error types. In contrast, model-based methods rely on explicit heuristic functions~\cite{liu2025robodexvlm} or sensors of the robot~\cite{parastegari2018failure} to detect discrepancies between expected and observed outcomes, often by monitoring execution constraints or key performance indicators and triggering corrective actions when violations occur.
Both paradigms have limitations: learning-based methods can struggle to generalize to unseen failure modes, while model-based pipelines may be computationally expensive and may not capture complex contact dynamics. These limitations motivate our reflective recovery mechanism, which leverages VLM-based semantic verification~\cite{singh2025malmm} together with structured memory to support diagnosis and replanning during long-horizon or complicated execution.

\section{Embodied Agentic Structure for Robotic Manipulation}
\label{sec:agent_manipulation}

In this section, we formulate the problem of robotic manipulation and detail the architecture of our proposed embodied AI agent. We focus on the semantic parsing of unstructured environments and the coordination mechanism that bridges high-level reasoning with low-level control.

\subsection{Problem Statement}
We formulate robotic manipulation as a sequential decision-making problem over a horizon $T$, where an agent selects actions $a_t \in \mathcal{A}$ at discrete time steps $t \in \{0,1,\dots,T-1\}$ to drive the environment from an initial state $s_0 \in \mathcal{S}$ toward a goal specification $s_g$ (or, more generally, a goal set $\mathcal{S}_g \subseteq \mathcal{S}$). The resulting trajectory is denoted by the sequence $\tau = (s_0, a_0, s_1, a_1, \dots, s_T)$, where state transitions are governed by the stochastic transition dynamics $P(s_{t+1} \mid s_t, a_t)$ of the underlying Markov Decision Process (MDP). The objective is to find a policy $\pi: \mathcal{S} \times \mathcal{S}_g \rightarrow \mathcal{A}$ that maximizes the probability of successfully and safely completing the task.

Specifically, long-horizon tasks exhibit strong \textit{sequential dependency}: failures or small deviations at early steps can invalidate later actions. If we define an atomic-step success event $E_t$, e.g., satisfying a pre-/post-condition for $a_t$, the episode success probability can be expressed as
\begin{equation}
P(\text{success}) = P\Big(\bigcap_{t=0}^{T-1} E_t\Big) = \prod_{t=0}^{T-1} P\big(E_t \mid E_{0:t-1}\big),
\end{equation}
which typically decreases as the horizon grows. This motivates explicit verification and recovery mechanisms to mitigate error accumulation during execution.

The system utilizes a constrained perception suite and free-form textual commands to facilitate universal deployability across diverse, unstructured environments. The robot perceives the workspace using a single wrist-mounted RGB-D camera, producing an ego-centric observation $\mathcal{I}_w \in \mathbb{R}^{H \times W \times 4}$. We deliberately exclude static third-person views to avoid external instrumentation and to support mobile deployment.
Furthermore, the task is specified by a free-form natural language command $\mathcal{L}_h$, which must be semantically grounded into executable operations. The control output is defined in the end-effector (EEF) Cartesian space: an action $a_t$ specifies a target EEF pose, represented as $\mathbf{p}_t = [x, y, z, q_w, q_x, q_y, q_z]^\top \in \mathbb{R}^7$, where $[x,y,z]$ denotes position and $[q_w,q_x,q_y,q_z]$ denotes a unit quaternion for orientation. 

\subsection{Agentic AI Coordinated Workflow}
To address the complexity of long-horizon tasks, we propose a hierarchical agentic workflow that decouples semantic reasoning from geometric execution. We first define the structured world model used for grounding and then describe how the agent plans and executes long-horizon behaviors on top of it.

\subsubsection{Semantic-Operational State Graph}
The SOSG serves as the structured world model for our manipulation framework, as illustrated in the bottom panel of Fig.~\ref{fig:aog_structure}. We denote the graph as $\mathcal{G} = (\mathcal{V}, \mathcal{E})$. The node set $\mathcal{V} = \{v_i\}_{i=1}^n$ represents physical entities in the workspace. Each node $v_i$ is parameterized by an augmented state vector $\mathbf{x}_i \in \mathcal{X}$ that encapsulates geometric, kinematic, and semantic properties needed for motion synthesis:
\begin{equation}
    \mathbf{x}_i = \left[ \mathcal{S}_i, \mathcal{K}_i, \mathcal{A}_i, \mathcal{P}_i \right]^\top.
\end{equation}
The components are defined as follows:
\begin{itemize}
    \item $\mathcal{S}_i = \{l_i, r_i, \tau_i\}$ represents the \textbf{semantic configuration}, comprising the object class $l_i$, its operational role $r_i$ (e.g., container, manipulatable object), and a natural language context $\tau_i$ for high-level task planning.
    \item $\mathcal{K}_i = \{\alpha_i, w_i, \mathcal{R}_i\}$ defines the \textbf{kinematic and spatial constraints}, where $\alpha_i \in \{0, 1\}$ indicates whether the object is articulated (i.e., possessing internal degrees of freedom such as prismatic joints), $w_i \in \mathbb{R}^+$ specifies a characteristic grasping aperture for stable prehension, and $\mathcal{R}_i$ denotes the pose relative to the robot base frame $\mathcal{F}_b$.
    \item $\mathcal{A}_i = \{c_i, \zeta_i, m_i\}$ captures \textbf{intrinsic physical attributes}, including color $c_i$, geometry/shape $\zeta_i$, and material properties $m_i$, which inform contact and interaction strategies.
    \item $\mathcal{P}_i = \{ (p_k, \delta_k) \}_{k=1}^K$ defines the \textbf{operational part decomposition}, mapping sub-components $p_k$ to functional descriptions $\delta_k$ for part-centric manipulation (e.g., grasping a drawer handle).
\end{itemize}
By embedding these physical and semantic priors into $\mathbf{x}_i$, the SOSG provides a unified representation that bridges high-level symbolic planning with low-level position control. \textcolor{lblue}{An SOSG edge $e_{ij}=(v_i,\rho_{ij},v_j)$ is instantiated only when perception or an executed tool provides evidence for a relation $\rho_{ij}$, where $\rho_{ij}$ is an object--object spatial relation, a part--whole relation, or an operation-grounding relation. After operation $o_k$, the verifier returns $(E_t,\hat{\mathcal{G}}_t)$; successful verification commits $\hat{\mathcal{G}}_t$ as the next episodic state, whereas failure records the diagnostic label and marks only the affected nodes/edges as stale before local repair or global replanning. The schema is intentionally compact and task-driven: class, role, pose, articulation, grasp aperture, and functional parts are populated on demand, rather than assuming a complete manually annotated world model.}

\subsubsection{Agentic Task and Action Planning}
\begin{figure}[t]
    \centering
    \includegraphics[width=1\linewidth]{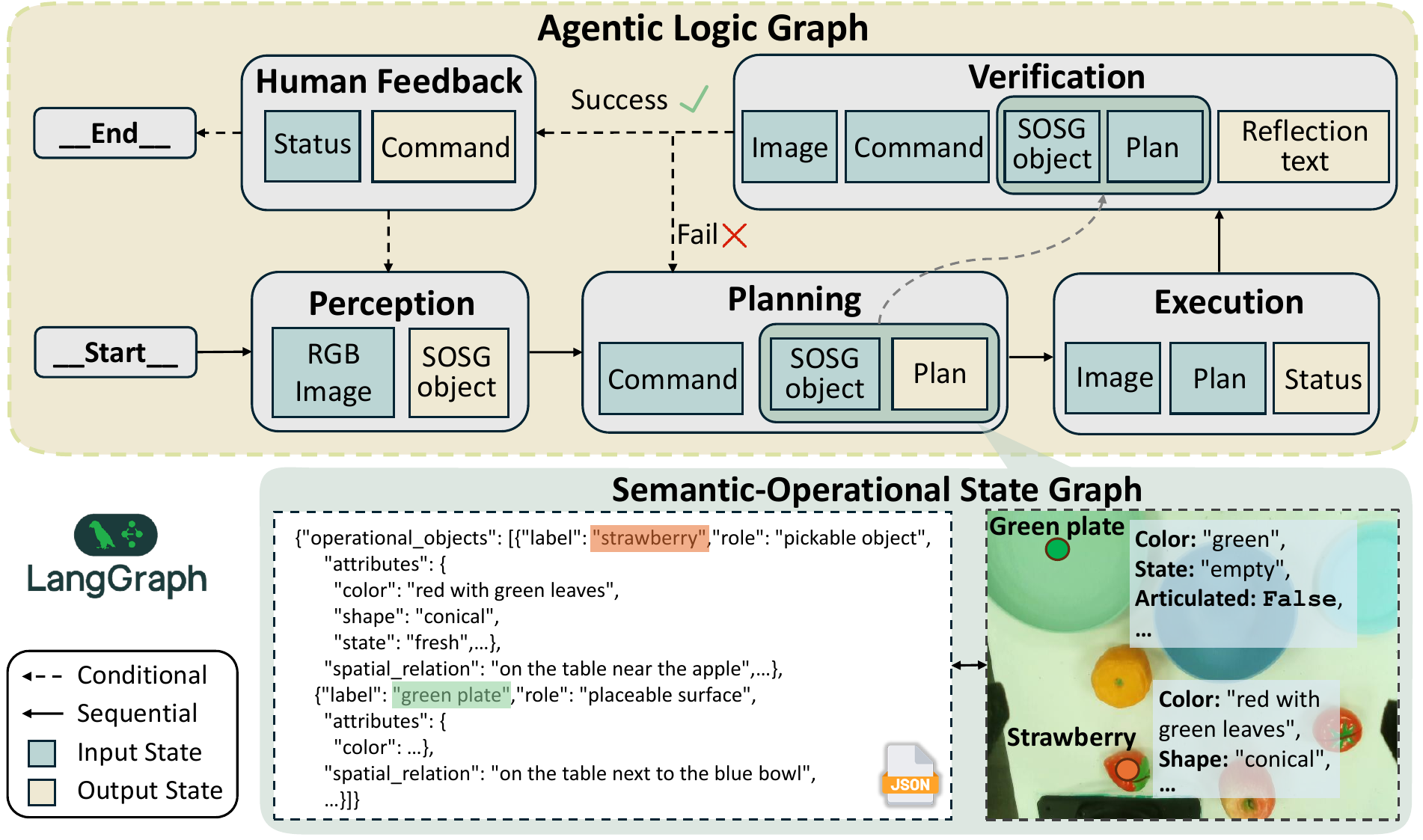}
    \caption{\textcolor{lblue}{Bidirectional coupling between the Agentic Logic Graph (ALG) and Semantic-Operational Scene Graph (SOSG). ALG operation nodes query SOSG geometry, affordances, and verified preconditions to ground tool parameters; execution verification writes outcomes and updated states back to SOSG, enabling local repair or global replanning.}}
    \label{fig:aog_structure}
    \vspace{-0.3cm}
\end{figure}

Building on the SOSG, the ALG follows a \textit{Perceive-Plan-Act-Reflect} cycle, as illustrated in the top panel of Fig.~\ref{fig:aog_structure}. Given a natural language command $\mathcal{L}_h$ and the available tool library $\mathcal{T}_c$, the agent first perceives the scene and instantiates/updates $\mathcal{G}$. It then decomposes the long-horizon objective into a sequence of executable sub-tasks and executes each sub-task by invoking modules for spatial reasoning, affordance generation, and collision-free motion planning.
The toolset $\mathcal{T}_c$ is conditioned on the robot configuration $c \in \{\mathtt{fixed}, \mathtt{mobile}\}$. Specifically, we define $|\mathcal{T}_{\mathtt{fixed}}| = 2$ and $|\mathcal{T}_{\mathtt{mobile}}| = 3$, reflecting the varying operational capabilities of each platform, detailed in Section~\ref{subsec:tools}. Crucially, the workflow maintains a reflection loop: if a sub-task fails, the agent retrieves recent interaction history as the episodic memory to diagnose the cause and replan, rather than halting.

\subsubsection{Agentic Spatial Reasoning and Action Parameterization}
To translate high-level intent into physical execution, the agent performs fine-grained spatial reasoning by grounding the command $\mathcal{L}_h$ into specific nodes within the SOSG. This process identifies task-relevant objects to compute the affordance $\mathcal{A}_{ff}$ and its corresponding 7-DoF grasp/release offset $\mathcal{J} = \{\bm t, \bm R, w\}$, where $\bm t \in \mathbb{R}^3$ and $\bm R \in SO(3)$ represent the translation and orientation, and $w$ denotes the target gripper aperture. This parameterization bridges the gap between semantic intent and the geometric constraints of the workspace.
Once $\mathcal{J}$ is determined, the agent coordinates the motion planning function $f_{plan}$ to generate an executable trajectory:
\begin{equation}
    \mathcal{W} = f_{plan}(\mathbf{p}_{s}, \mathbf{p}_{t}, \mathcal{M}_{occ})
\end{equation}
where $\mathbf{p}_{s}$ and $\mathbf{p}_{t}$ are the starting and target EEF poses, and $\mathcal{M}_{occ}$ is the occupancy representation derived from RGB-D inputs. The output $\mathcal{W} = \{\mathbf{w}_1, \mathbf{w}_2, \dots, \mathbf{w}_k\}, \mathbf{w}_i\in \mathbb{R}^7,$ is a sequence of collision-free waypoints that ensures safe navigation through the environment. By delegating trajectory generation to this module, the agent abstracts low-level obstacle avoidance from its high-level reasoning flow. Specific implementation details regarding the planner and inverse kinematics are provided in Section \ref{sec:action_generation}.

\section{Integrated Agentic Graph Execution for Robotic Manipulation}
\label{sec:action_generation}

This section details how UniManip realizes a complete agentic operational graph.
By leveraging the ALG as procedural memory and the SOSG as grounded episodic memory, the framework systematically achieves grounded scene parsing, action abstraction, safety-aware motion synthesis, and closed-loop reflection. It ensures that the robot remains synchronized with the physical environment even as tasks increase in complexity.

\subsection{Graph-Grounded Scene Parsing and Task Decomposition}
\subsubsection{Task Decomposition}\label{subsubsec:taskdecomp}
The task planning module bridges the free-form command $\mathcal{L}_{h}$ with the agent's episodic world model $\mathcal{G}=(\mathcal{V},\mathcal{E})$. Concretely, the ALG maps $\mathcal{L}_{h}$ and the available tool library $\mathcal{T}_c$ into a sequence of graph-conditioned sub-tasks, each represented as an operation over nodes (and parts) in $\mathcal{V} \cup \left( \bigcup_{i=1}^{n} \mathcal{P}_i \right)$.

A sub-task is encoded as either a directed edge $e_{ij}$ or a self-loop $e_{ii}$ with associated continuous parameters:
\textbf{Transit operations}, illustrated as the first tool in the bottom left panel of Fig.~\ref{fig:overview}, are represented by directed edges $e_{ij}$ and correspond to a \textit{move\_to} operation, $\mathbf{M}(v_i, \mathbf{g})$, which moves the end-effector from the state associated with $v_i$ to a target configuration $\mathbf{g}$ grounded on $v_j$.
\textbf{Manipulation operations}, shown as the second tool in Fig.~\ref{fig:overview}, are represented by self-loops $e_{ii}$ and correspond to \textit{operate} actuation, $\mathbf{O}(\mathbf{t}, \mathbf{r})$, where
\begin{equation}\label{eq:spatial_description}
\left\{
\begin{aligned}
\mathbf{t} &= [\delta_x, \delta_y, \delta_z]^\top, && \text{w.r.t. } \mathcal{F}_i \\
\mathbf{r} &= [\delta_{roll}, \delta_{pitch}, \delta_{yaw}]^\top, && \text{w.r.t. } \mathcal{F}_k
\end{aligned}
\right.
\end{equation}
denote rotation and translation values. Note that $\mathcal{F}_i,\mathcal{F}_j\in \{\mathcal{F}_b,\mathcal{F}_e\}$, where $\mathcal{F}_e$ is the EEF frame. This operation drives the EEF to execute relative motions along specified axes of the assigned Cartesian frame.
For instance, for the task of opening a drawer, the corresponding motion representations are illustrated in Fig.~\ref{fig:spatialPipeline}. This graph-based motion representation makes the plan explicit, compositional, and amenable to verification and re-structuring during reflection.

\subsubsection{Dynamic SOSG Update and Scene Parsing}
To maintain consistency between the internal episodic world model and the physical environment, the SOSG is updated online during execution. The primary attribute requiring real-time refinement is the object pose $\mathcal{R}_i$ (and, when applicable, part states in $\mathcal{P}_i$). We adopt a hierarchical open-vocabulary perception pipeline to update these attributes:
\textbf{Coarse retrieval:} given the natural language context $\tau_i$ in $\mathcal{S}_i$ and the current observation $\mathcal{I}_w$, an open-vocabulary detector proposes candidate regions $\{\mathbf{B}_k\}$ aligned with $\tau_i$~\cite{liu2024grounding}. Candidates exceeding a confidence threshold $\tau_d$ are retained as region proposals $\mathbf{B}^*$.
\textbf{Fine refinement:} a promptable segmentation model refines $\mathbf{B}^*$ into a pixel-accurate mask $\mathcal{M}_{seg}$~\cite{ravi2025sam}. Together with the synchronized depth channel in $\mathcal{I}_w$, this mask enables robust 3D pose estimation and grasp perception. Concurrently, this unified pipeline facilitates affordance detection, allowing the updated 3D pose $\mathcal{R}_i$ of the target object to be seamlessly integrated into the SOSG. This approach ensures robust state tracking even in the presence of clutter and partial occlusion.

For articulated objects ($\alpha_i=1$), the agent additionally maintains an episodic mechanism state over operational parts $\{\delta_k\}_{k=1}^K$ in $\mathcal{P}_i$ (e.g., \textit{upper drawer: open}). This state is refreshed after each interaction and provides the agentic logic layer with grounded, temporally consistent pre-/post-conditions for subsequent planning.
\subsection{Action Abstraction through Agentic Operational Primitives}\label{subsec:tools}
\begin{figure}[t]
    \centering
    \includegraphics[width=0.77\linewidth]{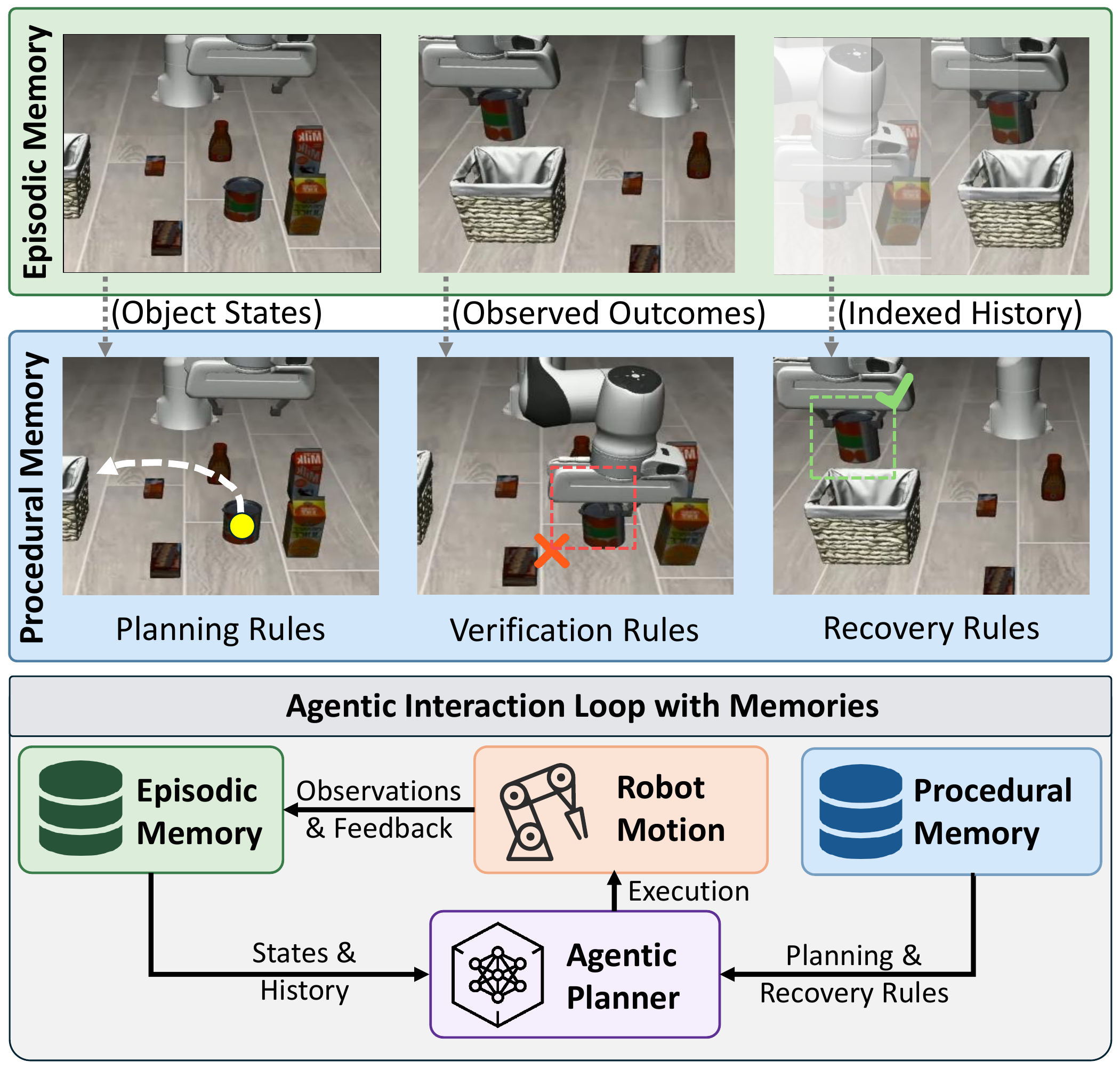}
    \caption{\textcolor{lblue}{Dual-memory mechanism. Episodic Memory stores task history, object states, action outcomes, and failure diagnostics for recovery. Procedural Memory stores executable skill primitives, affordance graphs, tool-selection logic, and recovery branches. Procedural Memory reads the current episodic state to select the next operation, while verification updates Episodic Memory after every execution.}}
    \label{fig:memory_illustration}
\end{figure}
\begin{figure}[t]
    \centering
    \includegraphics[width=1\linewidth]{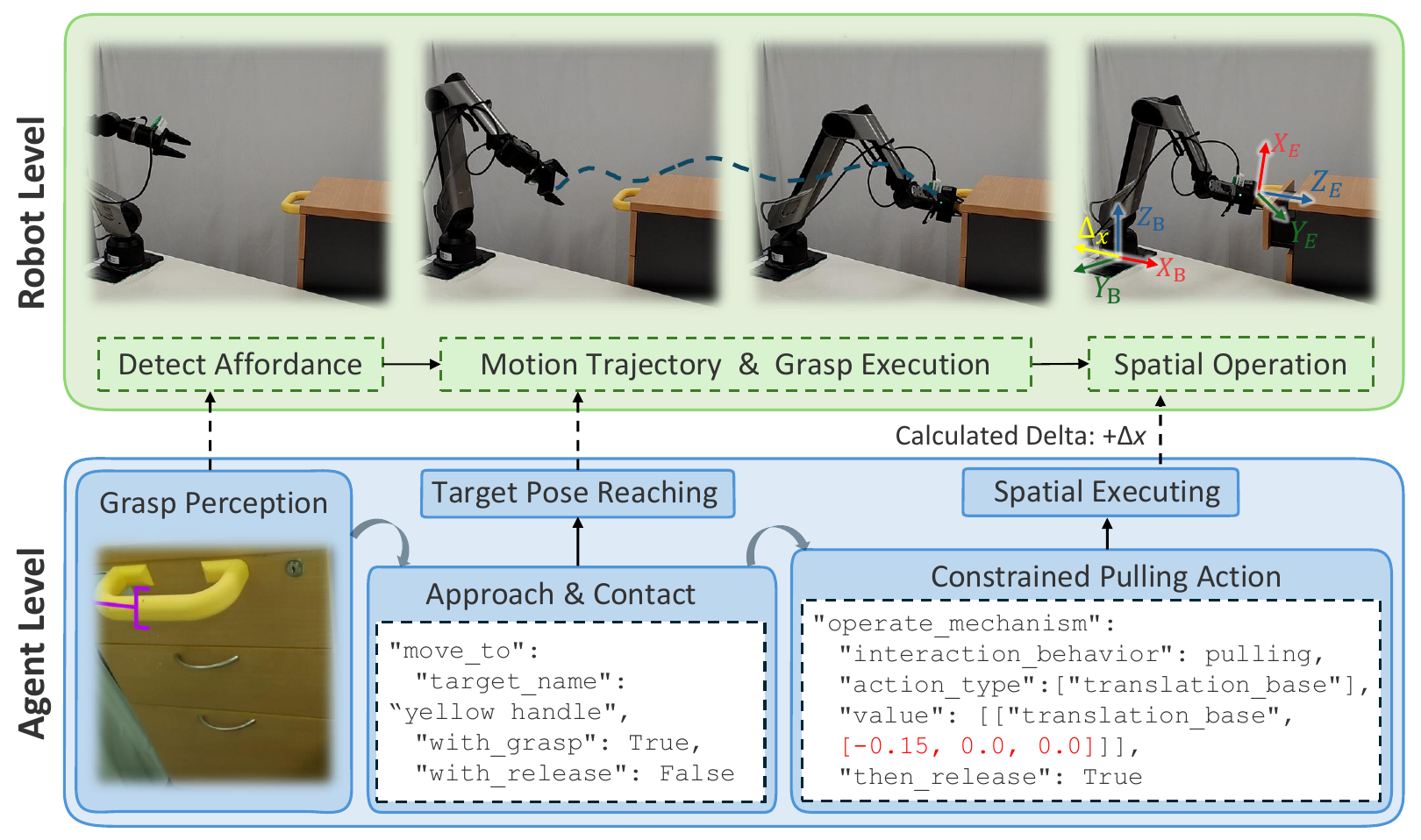}
    \caption{Demonstration of the spatial operations of the robot, with an instance of opening a drawer. The task is decomposed into several tool invocations, and each tool has its specific spatial operational formats for the movement of the robotic manipulator.}
    \label{fig:spatialPipeline}
\end{figure}
The agentic graph executes the planned sub-tasks by invoking tools from the primitive library $\mathcal{T}_c$ that implements atomic spatial operations under clear geometric semantics~\cite{carneiro2002herbert}. Each tool takes structured inputs grounded on the current SOSG $\mathcal{G}_\tau$ and returns parameters that can be tracked by the low-level controller.
In the fixed-base setting, 
\begin{equation}
    \mathcal{T}_c=\{\mathbf{M}(v_i, \mathbf{g}),\mathbf{O}(\mathbf{t}, \mathbf{r}),\cdots\}
\end{equation}
primarily consists of \textit{objective-oriented} motion and \textit{operation-oriented} motion, as referred in Section~\ref{subsubsec:taskdecomp}. This separation allows the agentic logic layer to compose long-horizon behavior as sequences of \emph{goal-directed} and \emph{constraint-directed} spatial operations while keeping the execution consistent.

Without loss of generality, the same abstraction extends to other robot morphologies by adding embodiment-specific tools without changing the task-level representation. Specifically, a mobile manipulator needs an augmented library $\mathcal{T}_\mathtt{mobile}$ with a navigation primitive
\begin{equation}
    \mathbf{N}(\mathcal{O}_{target}, \hat{d}_{obj}, d_{goal})\in \mathcal{T}_\mathtt{mobile},
\end{equation}
where $\mathcal{O}_{target}$ denotes the target entity to dock to, $\hat{d}_{obj}$ is the estimated real-time distance from the base to the target (obtained online from the ego-centric observation), and $d_{goal}$ is the desired docking clearance. This primitive drives the base to a manipulation-feasible configuration, after which the same manipulation tools in $\mathcal{T}_c$ can be invoked with unchanged semantics.
The specific tools are detailed as follows.
\subsubsection{Objective-Oriented Motion}
This tool directs the EEF to a spatial target grounded on the current SOSG $\mathcal{G}_\tau$. Its inputs include a target pose $\mathcal{R}_{target} \in SE(3)$ associated with a node $v_i \in \mathcal{V}$, together with boolean flags $b_{grasp}, b_{release} \in \{0,1\}$ indicating whether a gripper action should be triggered upon arrival. The tool invokes the collision-free motion planner (Section~\ref{subsec:collision_free}) to generate a feasible trajectory, and outputs a trackable trajectory segment $\mathcal{T}_{seg}$ executed by the low-level advanced controller.

\subsubsection{Operation-Oriented Motion}
This tool executes precise relative motions specified by metric constraints, which is essential for interaction-centric tasks such as pushing, pulling, twisting, or wiping. The input consists of a reference frame $\in \{\mathcal{F}_b, \mathcal{F}_e\}$ (base frame or end-effector frame) and a 6-DoF motion increment denoted in (\ref{eq:spatial_description}), which encodes translation and rotation along the axes of specific frames. Let $T_{curr} \in SE(3)$ denote the current EEF pose. We construct an incremental transform $\Delta T(\boldsymbol{\delta}) \in SE(3)$ based on the operation values in (\ref{eq:spatial_description}) and compute the target pose by left- or right-multiplication depending on the chosen reference frame:

\begin{equation}
\left\{
\begin{aligned}
T_{new} &= \Delta T(\mathbf{t,r})\, T_{curr}, && \text{w.r.t. } \mathcal{F}_b \\
T_{new} &= T_{curr}\, \Delta T(\mathbf{t,r}), && \text{w.r.t. } \mathcal{F}_e
\end{aligned}
\right.
\end{equation}
This formulation provides a uniform interface for part-centric operations (e.g., pulling a drawer along an axis) while remaining compatible with downstream trajectory generation and closed-loop verification.

\subsection{Agentic-Guided Motion Synthesis}
\label{subsec:collision_free}

Following the operational primitives, the safety-aware motion generation maps a target EEF pose to a collision-free waypoint sequence in Cartesian coordinate $\mathcal{F}_b$. Our design targets \emph{single-view} deployment and therefore explicitly accounts for occlusion-induced uncertainty by constructing a conservative free space.

\subsubsection{Defensive Reconstruction of Occluded Areas}
Given a single RGB-D observation $\mathcal{I}_w$, we first reconstruct a conservative volumetric occupancy map that mitigates the \textit{floating obstacle} issue caused by self-occlusions (e.g., unobserved cavities behind object rims). Let $\mathcal{P}_{raw}$ denote the raw point cloud back-projected from $\mathcal{I}_w$. We discretize the workspace into a voxel grid with resolution $r$ (meters) and define an occupancy map
\begin{equation}
\begin{split}
    \mathcal{M}_{init}: & \{0,\dots,H-1\} \times \{0,\dots,W-1\} \\
    & \times \{0,\dots,D-1\} \to \{0,1\},
\end{split}
\end{equation}
where a voxel indexed by $\mathbf{u}=[u,v,w]^\top$ is marked occupied if any point in $\mathcal{P}_{raw}$ falls within its spatial cell. In all experiments we use $r=0.01\,\mathrm{m}$ to balance accuracy and efficiency.

\begin{figure}[t]
    \centering
    \includegraphics[width=1\linewidth]{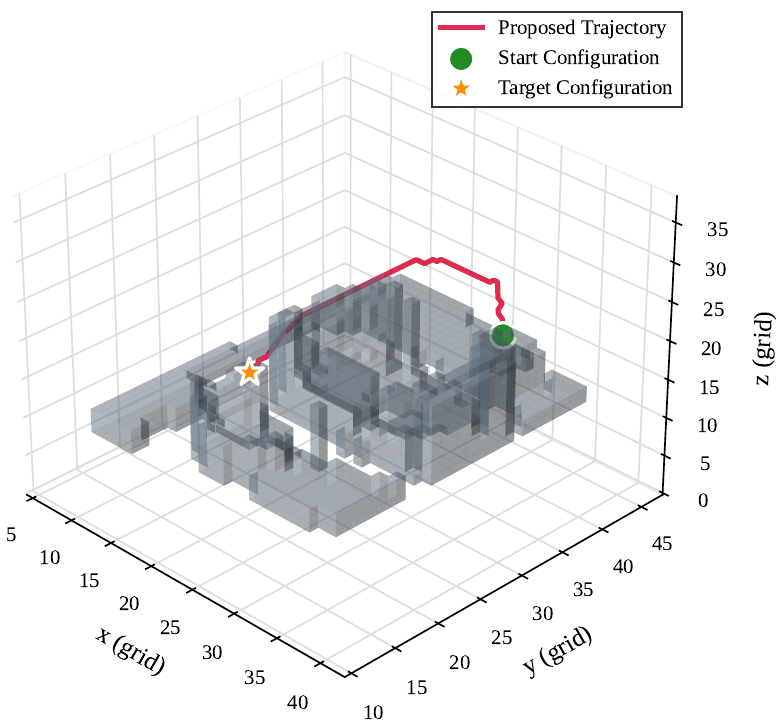}
    \caption{Visualization of the conservative volumetric occupancy grid $\mathcal{M}_{final}$ generated from a single-view RGB-D observation. The gravity-aligned completion over-approximates unknown space, improving safety under occlusion.}
    \label{fig:occupancy_map}
\end{figure}

To reduce sensor noise and fill small holes on observed surfaces, we apply 3D morphological closing with a structuring element (kernel) $\mathcal{K}$,
\begin{equation}
    \mathcal{M}_{closed} = \mathcal{M}_{init} \bullet \mathcal{K} = (\mathcal{M}_{init} \oplus \mathcal{K}) \ominus \mathcal{K},
\end{equation}
where $\oplus$ and $\ominus$ denote dilation and erosion, respectively. 
For manipulation in tabletop and indoor scenes, we adopt a conservative safety heuristic that enforces vertical support under occupied voxels. Specifically, if a voxel is occupied, all voxels below it along the gravity axis are treated as non-traversable. The core objective is to constrain the planner from traversing unobserved cavities where occupancy is probable, such as the occluded space beneath a basket's rim, thereby mitigating collision risks in partially mapped environments.
The final conservative map is
\begin{equation}
    \mathcal{M}_{final}(u,v,w) = \bigvee_{k=w}^{D-1} \mathcal{M}_{closed}(u,v,k),
\end{equation}
where $\bigvee$ is the logical \textit{or}. Fig.~\ref{fig:occupancy_map} visualizes an optimized trajectory example along with the reconstructed occupancy grid in the scenario with an obstacle blocking the EEF from the object to the target container, shown in Fig.~\ref{fig:raw_scenario_grid}.

\subsubsection{Safety-Aware Trajectory Generation}
\begin{figure}[t]
    \centering
    \subfigure[Corresponding Scenario]{
        \includegraphics[width=0.80\linewidth]{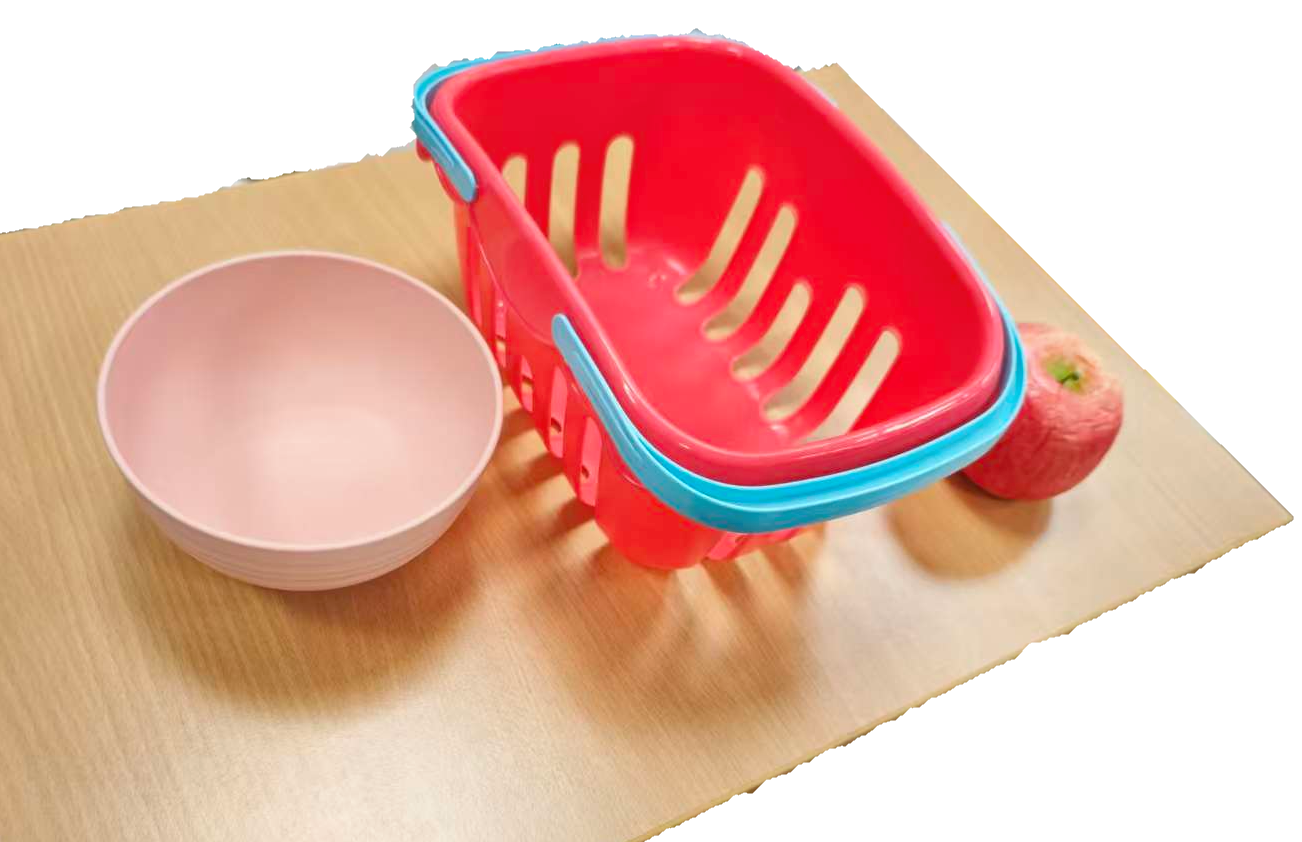}
        \label{fig:raw_scenario_grid}
    }
    \hfill
    \subfigure[Sliced ESDF Heat Map]{
        \includegraphics[width=0.95\linewidth]{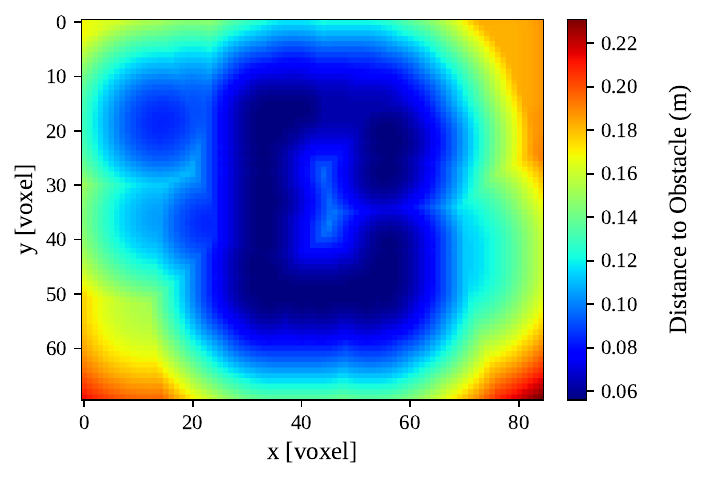}
        \label{fig:esdf_slice}
    }
    \caption{The corresponding scene and a sliced ESDF illustrating clearance values from obstacles. The ESDF is derived from the conservative occupancy grid, which accounts for occlusions and ensures safe trajectory planning.}
    \label{fig:outdoor_landing_visuals}
\end{figure}

We convert the conservative occupancy $\mathcal{M}_{final}$ into an Euclidean signed distance field (ESDF) $\Phi$, where each grid location is assigned its distance to the nearest occupied voxel:
\begin{equation}
    \Phi(\mathbf{u}) = \operatorname*{dist}(\mathbf{u}, \{\mathbf{u}'\mid \mathcal{M}_{final}(\mathbf{u}')=1\}).
\end{equation}
This representation induces a smooth clearance metric that can be queried efficiently during planning.
Let $\mathbf{u}_s$ and $\mathbf{u}_g$ denote the start and goal grid indices corresponding to the start and target EEF poses. We compute a discrete path $\mathcal{P}^* = [\mathbf{u}_0,\dots,\mathbf{u}_L]$ using A* algorithm~\cite{liu2025udmc} with objective:
\begin{equation}
    f(\mathbf{u}) = g(\mathbf{u}) + h(\mathbf{u}),
\end{equation}
where $g$ is the accumulated edge cost and $h$ is an admissible heuristic to $\mathbf{u}_g$. Specifically, Fig.~\ref{fig:esdf_slice} illustrates a sliced ESDF heat map derived from the conservative occupancy grid, highlighting clearance values from obstacles in the environment.
Note that we enforce collision avoidance by requiring a clearance $r_\text{safe}>\SI{0.05}{m}$ at each expanded node,
\begin{equation}
    V(\mathbf{u}) = \mathbb{I}\big[\Phi(\mathbf{u}) \ge r_\text{safe}\big],
\end{equation}
where $\mathbb{I}[\cdot]$ is the indicator function. 
This EEF trajectory generation approach is equivalent to planning in a free space eroded by a ball of radius $r_\text{safe}$ (i.e., a conservative configuration-space dilation) without explicitly constructing the Minkowski sum.

The discrete path $\mathcal{P}^*$ is mapped to metric coordinates in the base frame $\mathcal{F}_b$ as $\mathbf{p}_i = r \cdot \mathbf{u}_i + \mathbf{p}_{\text{origin}}$, for $i=0,\dots,L$, where $\mathbf{p}_{\text{origin}}$ denotes the world coordinates of the voxel grid origin. To extend the discrete geometric path $\{\mathbf{p}_i\}$, initially defined only by Cartesian positions, into a full 6-DOF end-effector trajectory, we perform orientation synthesis via Spherical Linear Interpolation (Slerp)~\cite{shoemake1985animating}. 
Let $\boldsymbol{\xi}_s, \boldsymbol{\xi}_g \in S^3$ denote the unit quaternions representing the initial and goal orientations, respectively. For each waypoint $\mathbf{u}_i$, we define the interpolation parameter $\alpha = i/L$. The synthesized orientation $\boldsymbol{\xi}_i$ is computed to traverse the shortest geodesic arc on the unit hypersphere:
\begin{equation}
\text{Slerp}(\boldsymbol{\xi}_s, \boldsymbol{\xi}_g; \alpha) = \frac{\sin((1-\alpha)\theta)}{\sin\theta}\boldsymbol{\xi}_s + \frac{\sin(\alpha\theta)}{\sin\theta}\boldsymbol{\xi}_g,
\end{equation}
where $\theta = \arccos(\boldsymbol{\xi}_s \cdot \boldsymbol{\xi}_g)$ is the angle between the orientations. To ensure the shortest path is taken on $S^3$, we enforce the condition $\boldsymbol{\xi}_s \cdot \boldsymbol{\xi}_g \geq 0$, negating $\boldsymbol{\xi}_g$ if necessary to account for the double-cover property of $SO(3)$. 
This synthesis produces a trajectory that exhibits smooth rotational transitions while preserving the safety clearance established by the ESDF.

\subsubsection{Relaxed Inverse Kinematics for Waypoint Tracking}
While the collision-free motion planner outputs a geometric waypoint sequence $\mathcal{W} = \{\mathbf{w}_1, \dots, \mathbf{w}_k\}$ in task space, executing this path requires solving inverse kinematics to map each desired end-effector pose to a joint configuration $\mathbf{q} \in \mathbb{R}^n$. In practice, standard IK solvers can fail when a waypoint is close to a kinematic singularity, lies marginally outside the reachable workspace, or requires an orientation that is difficult to realize under joint limits. Because the waypoint generator is agnostic to the arm's kinematic feasibility set, such failures can halt long-horizon execution even when a nearby pose would be sufficient for safe progress.

To address this issue, we adopt a \emph{parameter-relaxed} IK strategy that prioritizes continuity of execution over exact pose matching in kinematically challenging regions.
Let $\mathbf{p}(\mathbf{q}) \in \mathbb{R}^3$ and $\boldsymbol{\xi}(\mathbf{q}) \in S^3$ denote the forward kinematics position and unit quaternion orientation of the end-effector, respectively, where $\mathbf{q} \in \mathbb{R}^n$ is the joint angle vector. Given a desired pose, the optimal joint configuration $\mathbf{q}^*$ is obtained by solving:
\begin{equation}
\mathbf{q}^* = \operatorname*{arg\,min}_{\mathbf{q}}\; \lambda_p \left\lVert \mathbf{p}(\mathbf{q}) - \mathbf{p}_{\text{des}} \right\rVert_2^2 + \lambda_r \left\lVert \boldsymbol{\xi}(\mathbf{q}) \ominus \boldsymbol{\xi}_{\text{des}} \right\rVert_2^2
\end{equation}
subject to joint-limit constraints:
\begin{equation}
\mathbf{q}_{\min} \preceq \mathbf{q} \preceq \mathbf{q}_{\max},
\end{equation}
and nominal convergence tolerances:
\begin{equation}
\left\lVert \mathbf{p}(\mathbf{q}) - \mathbf{p}_{\text{des}} \right\rVert_2 < \epsilon_p, \quad \left\lVert \boldsymbol{\xi}(\mathbf{q}) \ominus \boldsymbol{\xi}_{\text{des}} \right\rVert_2 < \epsilon_r.
\end{equation}
Here, $\lambda_p, \lambda_r > 0$ are weighting coefficients, and $\ominus$ denotes a geodesic orientation-error operator on $SO(3)$, typically implemented via the quaternion logarithm or angle-axis distance. Many conventional IK implementations keep $(\epsilon_p, \epsilon_r)$ fixed and tight (e.g., millimeter-level position tolerance), which can yield ``no-solution'' outcomes for waypoints that are only marginally infeasible due to voxelization or ESDF gradients.

We mitigate non-convexity via multi-seed initialization near the initial state and relax the tolerances only when necessary.
The solver first attempts convergence under strict tolerances $(\epsilon_p^{(0)},\epsilon_r^{(0)}) = (0.01\,\mathrm{m}, 0.02\,\mathrm{rad})$. If it fails to converge, we relax the tolerances according to
\begin{equation}
    \epsilon_p^{(k+1)} \leftarrow 5\,\epsilon_p^{(k)}, \quad \epsilon_r^{(k+1)} \leftarrow 2\,\epsilon_r^{(k)}.
\end{equation}
The loop terminates once a feasible solution is found or once $(\epsilon_p,\epsilon_r)$ exceeds a predefined safety budget.
To ensure a well-defined orientation error, we enforce quaternion normalization on the target, $\lVert \mathbf{q}_{\mathrm{des}} \rVert_2 = 1$, before optimization. Overall, RIK yields a best-effort joint solution that preserves smooth execution and avoids dead-ends when the ideal waypoint is near the boundary of kinematic feasibility. This is particularly beneficial for different embodiments with varying kinematic structures and reachability profiles.

\subsection{Agentic Verification and Autonomous Recovery}
\begin{figure}[t]
    \centering
    \includegraphics[width=\linewidth]{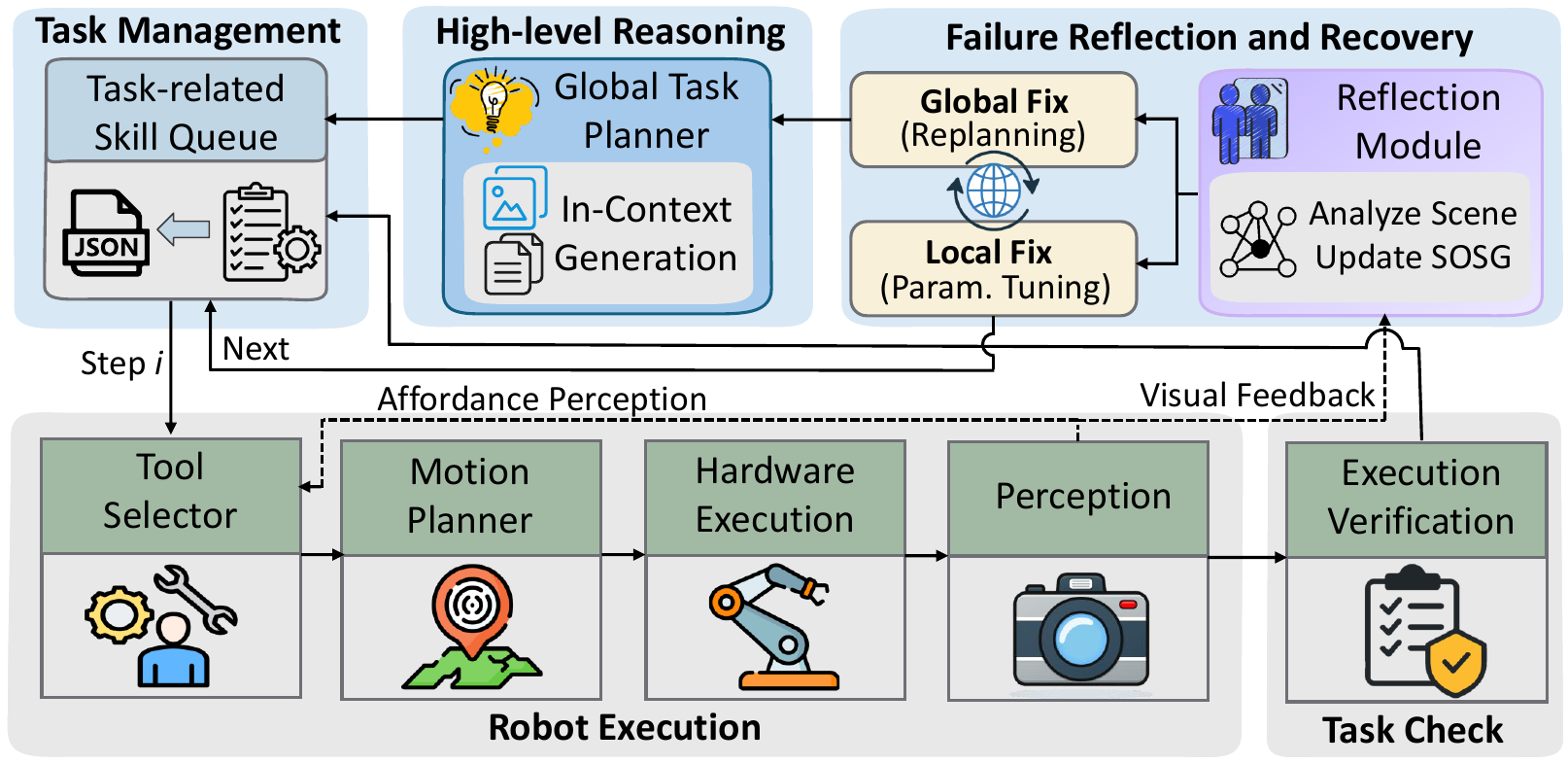}
    \caption{The closed-loop failure reflection and recovery mechanism in UniManip. The system continuously monitors execution states, detects anomalies, and invokes high-level cognitive reasoning to diagnose and recover from failures, either through local corrections or global replanning.}
    \label{fig:ai_agent_logic}
\end{figure}
A core limitation of open-loop task execution is its fragility under partial observability and unmodeled contact dynamics: a single failed atomic operation can invalidate subsequent preconditions and cascade into long-horizon failure. 
UniManip addresses this by embedding execution into a closed-loop failure reflection and recovery mechanism. It couples procedural memory (deciding what to do, when to verify, and how to replan) in the \textit{Tool Selector} of the agentic logic layer with the episodic memory (grounding the agent in object-centric, time-indexed world state) that functions in the Task Manager with SOSG $\mathcal{G}$. In this view, a failure is not a terminal outcome but a transitional state that triggers verification, diagnosis, and recovery actions conditioned on both the current observation and the structured execution trace stored in the AOG.

\subsubsection{Anomaly Detection and State Verification}
We represent a manipulation task $\mathcal{T}$ (parsed from the free-form instruction $\mathcal{L}_h$) as an ordered sequence of atomic spatial operations
\begin{equation}
    \mathcal{P}_{\mathrm{ops}} = (o_1, o_2, \dots, o_N).
\end{equation}
Let $s_t$ denote the robot-environment execution state at time step $t$ (including proprioception and the current episodic memory $\mathcal{G}_t$), and let $\mathcal{I}_t$ denote the corresponding wrist RGB-D observation. We introduce a verification operator $\mathcal{F}_{\mathrm{verify}}(s_t, \mathcal{I}_t, o_k)$ that evaluates whether the intended post-conditions of the current operation $o_k$ are satisfied.

Concretely, verification combines two complementary modalities. \textit{Geometric consistency} checks whether the measured robot state (e.g., end-effector pose, gripper state) satisfies the metric constraints implied by $o_k$ (e.g., waypoint reachability, relative motion bounds). \textit{Visual-semantic consistency} uses a VLM-based verifier to assess the post-execution observation and determine whether the SOSG (episodic memory) has transitioned as expected (e.g., an object is grasped, moved, or a drawer changes state).

The binary execution outcome is then defined as
\begin{equation}
    E_t = \begin{cases}
    1, & \text{if } \mathcal{F}_{\mathrm{verify}}(s_t, \mathcal{I}_t, o_k) = \texttt{success}, \\
    0, & \text{otherwise}.
    \end{cases}
\end{equation}
If $E_t=0$, the agent transitions from open-loop execution to the reflective recovery loop.

\subsubsection{Agentic Reflection via Graph and Memory}
Upon a detected failure ($E_t=0$), the agent initiates a reflection cycle that explicitly leverages the AOG as structured procedural memory. Let $\mathcal{H}_t$ denote a short-term execution trace (e.g., the most recent operation nodes, tool inputs/outputs, and verification results), and let $\mathcal{M}_{\mathrm{ret}}$ denote retrieved experience from long-term memory via retrieval-augmented generation.
We model reflection as a transformation operator
\begin{equation}
    (\mathcal{G}_{t}^{\mathrm{ref}}, D_{\mathrm{err}}) = \Psi(\mathcal{I}_t, \mathcal{H}_t, \mathcal{M}_{\mathrm{ret}} \mid \text{Agent}),
\end{equation}
where $\Psi$ performs two coupled updates. First, it updates episodic memory by revising the SOSG state to $\mathcal{G}_{t}^{\mathrm{ref}}$ (e.g., annotating a node/part state as ``not grasped'' or ``drawer partially open''), ensuring that subsequent reasoning is grounded in the observed outcome rather than the intended one. Second, it produces a semantic diagnosis $D_{\mathrm{err}}$ (e.g., grasp slippage, collision, occlusion-induced mislocalization) that conditions the choice of recovery action in the agentic logic layer.

\subsubsection{Error-Informed Reflection and Agentic Recovery}
Conditioned on the diagnosis $D_{\mathrm{err}}$ and the reflected episodic state $\mathcal{G}_{t}^{\mathrm{ref}}$, the ALG selects a recovery strategy at one of two levels. Crucially, the system leverages the residual error of the failed trajectory, derived from the discrepancy between the intended post-condition and the observed state, as a feedback signal to guide the subsequent repair.

For low-level deviations, such as minor pose errors, gripper slippage, or transient occlusions, the agent computes a corrective adjustment $\delta o$ by back-projecting the execution error into the operation parameter space. The agent then reissues a repaired operation $o'_k = o_k \oplus \delta o$.
This mechanism uses the failure offset as an informative prior to adjust action parameters (e.g., increasing gripper force or shifting a grasp pose), preserving the high-level task structure while repairing the physical realization.

On the contrary, for high-level divergences that invalidate future preconditions, such as target displacement or scene topology changes, the execution error acts as a state-update trigger. The system discards the remaining plan and regenerates a new operation sequence conditioned on the updated episodic memory, the free-form command $\mathcal{L}_h$, and retrieved experience $\mathcal{M}_{\mathrm{ret}}$:
\begin{equation}
    \mathcal{P}_{\mathrm{new}} = \Pi(\mathcal{G}_{t}^{\mathrm{ref}}, \mathcal{L}_h, \mathcal{M}_{\mathrm{ret}}).
\end{equation}
By embedding the failure context directly into $\mathcal{G}_{t}^{\mathrm{ref}}$, the planner ensures that the new trajectory $\mathcal{P}_{\mathrm{new}}$ is physically consistent with the modified environment, preventing cyclic failures.

This recovery mechanism provides a structured procedural scaffold for the VLM to decide when to repair locally versus when to restructure the entire plan, ensuring long-horizon robustness through continuous feedback integration.

{\color{lblue}
To formalize the closed-loop synergy between the ALG and the SOSG, Algorithm~\ref{alg:closed_loop_execution} outlines the unified execution pipeline of UniManip. The framework grounds the language instruction into structured subgoals, synthesizes executable trajectories, and verifies each outcome. The integer $R_{\max}$ is the task-level recovery budget, while $r$ counts repair or replanning attempts.
}
\begin{algorithm}[t]
\caption{\textcolor{lblue}{Closed-Loop AOG Execution in UniManip}}
\label{alg:closed_loop_execution}
\begingroup
\color{lblue}
\begin{algorithmic}[1]
\Require $\mathcal{I}_0$, instruction $\mathcal{L}_h$, SOSG $\mathcal{G}_0$, primitives $\mathcal{T}$, limits $T_{\max}, R_{\max}$
\Ensure Trajectory $\tau$, status $\sigma \in \{\texttt{success}, \texttt{failure}\}$
\State Initialize $t\gets0$, $k\gets1$, $r\gets0$, $\tau\gets\varnothing$, $\mathcal{H}_0\gets\varnothing$
\State $\mathcal{P}_{\mathrm{ops}} = (o_1,\ldots,o_N)\gets\Pi(\mathcal{L}_h,\mathcal{G}_0,\mathcal{T})$ via ALG
\While{$k\leq |\mathcal{P}_{\mathrm{ops}}|$ \textbf{and} $t < T_{\max}$}
    \State Update $\mathcal{G}_t$ from incoming observation $\mathcal{I}_t$
    \If{\textbf{not} $\mathrm{PrecondSatisfied}(o_k,\mathcal{G}_t)$}
        \If{$r\geq R_{\max}$} \State \Return $(\tau,\texttt{failure})$ \EndIf
        \State $r\gets r+1$; $(\mathcal{G}_t^{\mathrm{ref}},D_{\mathrm{err}})\gets\Psi(\mathcal{I}_t,\mathcal{H}_t,\mathcal{M}_{\mathrm{ret}})$
        \State $\mathcal{P}_{\mathrm{ops}}\gets\Pi(\mathcal{G}_t^{\mathrm{ref}},\mathcal{L}_h,\mathcal{M}_{\mathrm{ret}})$; $\mathcal{G}_{t+1}\gets\mathcal{G}_t^{\mathrm{ref}}$; $k\gets1$; $t\gets t+1$
        \State \textbf{continue}
    \EndIf

    \State $\mathcal{T}_{\mathrm{seg}}\gets\mathrm{PlanMotion}(o_k,\mathcal{G}_t,\mathcal{I}_t,\mathcal{T})$
    \State Execute $\mathcal{T}_{\mathrm{seg}}$, append to $\tau$, observe $(s_{t+1},\mathcal{I}_{t+1})$
    \State $(E_t,\hat{\mathcal{G}}_t)\gets\mathcal{F}_{\mathrm{verify}}(s_{t+1},\mathcal{I}_{t+1},o_k)$; append to $\mathcal{H}_{t+1}$
    
    \If{$E_t=1$}
        \State $\mathcal{G}_{t+1}\gets\hat{\mathcal{G}}_t$; $k\gets k+1$; $t\gets t+1$
        \State \textbf{continue}
    \EndIf

    \If{$r\geq R_{\max}$} \State \Return $(\tau,\texttt{failure})$ \EndIf
    \State $r\gets r+1$; $(\mathcal{G}_t^{\mathrm{ref}},D_{\mathrm{err}})\gets\Psi(\mathcal{I}_{t+1},\mathcal{H}_{t+1},\mathcal{M}_{\mathrm{ret}})$
    \If{$D_{\mathrm{err}}=\texttt{local}$}
        \State $\delta o \gets \mathrm{ComputeDelta}(\mathcal{I}_{t+1},\mathcal{H}_{t+1},D_{\mathrm{err}})$
        \State $o_k\gets o_k\oplus\delta o$
    \Else
        \State $\mathcal{P}_{\mathrm{ops}}\gets\Pi(\mathcal{G}_t^{\mathrm{ref}},\mathcal{L}_h,\mathcal{M}_{\mathrm{ret}})$; $k\gets1$
    \EndIf
    \State $\mathcal{G}_{t+1}\gets\mathcal{G}_t^{\mathrm{ref}}$; $t\gets t+1$
\EndWhile
\State \Return $(\tau, k > |\mathcal{P}_{\mathrm{ops}}| \,?\, \texttt{success} : \texttt{failure})$
\end{algorithmic}
\endgroup
\end{algorithm}
{\color{lblue}
This formulation couples subgoal scheduling in the ALG with state updates in the SOSG. Verified primitives commit new relations and advance the operation index; failures consume the bounded recovery budget and trigger local residual correction or agentic replanning.
}

\section{Experimental Results and Analysis}
\label{sec:experiments}
\subsection{Evaluation Goals and Environmental Setup}
Our experiments evaluate UniManip along three axes: (1) zero-shot manipulation performance under unseen objects and distractors, (2) robustness in cluttered scenes compared with representative hierarchical open-vocabulary planners, and (3) long-horizon and interaction-rich tasks that stress verification and recovery. In addition, we report ablations that isolate the effects of collision-free planning, relaxed IK, and the reflection/recovery loop, and we demonstrate cross-embodiment transfer on a mobile manipulator without fine-tuning.

\begin{figure}[t]
    \centering
    \includegraphics[width=0.85\linewidth]{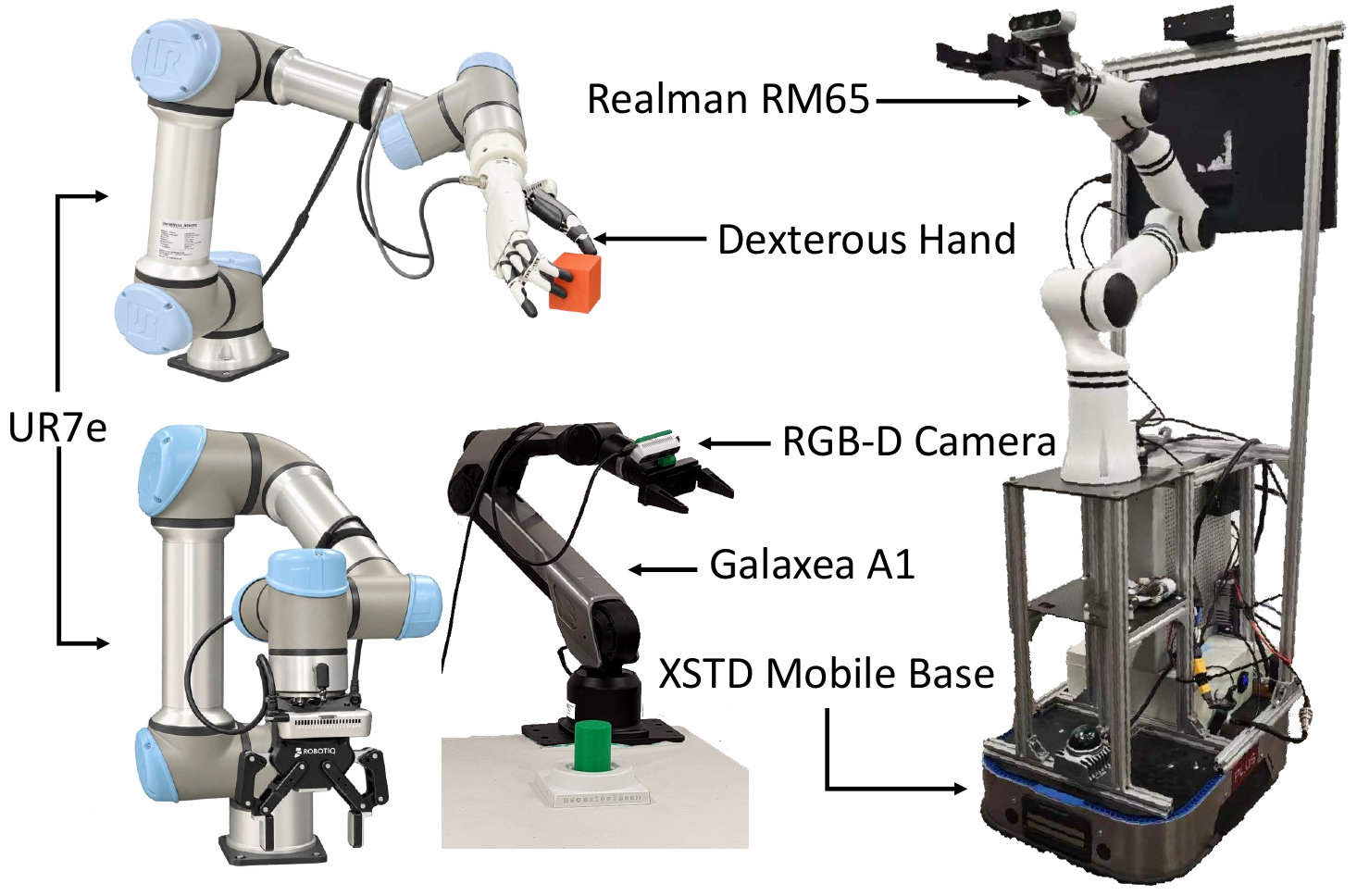}
    \caption{\textcolor{lblue}{\textbf{Physical experimental setups across diverse robotic embodiments.} The suite consists of the base Galaxea A1 arm with a wrist-mounted RGB-D camera, an XSTD mobile manipulator carrying a RealMan RM65 arm, and a UR7e arm interfaced with both a multi-fingered dexterous hand and a two-finger parallel gripper.}}
    \label{fig:robot_hardware}
\end{figure}

The experiments are conducted using a Galaxea A1 robotic arm (6-DoF) equipped with a two-finger gripper and an Intel RealSense D435 RGB-D camera mounted in an eye-in-hand configuration, as shown in Fig.~\ref{fig:robot_hardware}. The workspace is arranged with diverse household objects to evaluate zero-shot generalization across object categories, spatial layouts, and task formats.
All methods are executed on a workstation with an Intel i9-13900K CPU, an NVIDIA RTX 4080 GPU, and 32\,GB RAM. The software stack uses ROS Noetic with PyTorch for learning-based components and NVIDIA cuRobo for GPU-accelerated optimization.
For cross-embodiment evaluation, we additionally deploy UniManip on a mobile manipulator (right in Fig.~\ref{fig:robot_hardware}) consisting of a differential-drive base and a Realman RM65 (6-DoF) arm. The same eye-in-hand RGB-D sensing setup is used, enabling evaluation under viewpoint changes induced by navigation.

\textcolor{lblue}{The VLA baselines are fine-tuned from their released checkpoints on 1,000 Galaxea A1 teleoperation demonstrations, cross nine pick-and-place tasks, and use the wrist D435 plus a third-person D515 camera required by their input interface. Note that to align with pretraining distributions and target fine-tuning setups, modern VLA policies fundamentally require an external, static third-person camera. On the contrary, UniManip uses only the wrist camera and receives no task-specific demonstrations. ReKep, MOKA, VoxPoser, RoboEXP and UniManip w/o Graph are evaluated on the same task initializations and budget as UniManip. The latter uses LangChain to implement a minimal linear text plan with transient tool-call state in place of SOSG/ALG, while retaining the detector, segmentation, grasp generation, occupancy reconstruction, motion planner, IK solver, verifier, and recovery policy.}

\textcolor{lblue}{All tabletop comparisons conducted with the Galaxea A1 robot arm use 10 independently randomized trials per task and method; consequently, the original clutter benchmark includes 80 trials per method. A trial is considered successful only if the specified object reaches the designated receptacle without an emergency stop or manual intervention.}

\subsection{Zero-shot Manipulation Performance}
\label{sec:experiments_vla}
\begin{table*}[htbp]
  \centering
  \caption{Comparison with VLA Methods on Diverse Manipulation Tasks}
  \resizebox{\textwidth}{!}{%
    \begin{tabular}{cccccccccccccc}
    \toprule
    \multirow{2}[1]{*}{Task Format} & \multirow{2}[1]{*}{Task} & \multicolumn{3}{c}{$\pi_0$~\cite{black2024pi_0}} & \multicolumn{3}{c}{NORA~\cite{hung2025nora1}} & \multicolumn{3}{c}{NORA-1.5~\cite{hung2025nora}} & \multicolumn{3}{c}{UniManip (Ours)} \\
          &       & PSR$\uparrow$   & Dist.$\downarrow$ & SR$\uparrow$    & PSR$\uparrow$   & Dist.$\downarrow$ & SR$\uparrow$    & PSR$\uparrow$   & Dist.$\downarrow$ & SR$\uparrow$    & PSR$\uparrow$   & Dist.$\downarrow$ & SR$\uparrow$ \\
    \midrule
    \multirow{4}[2]{*}{Put \texttt{S} in \texttt{S}} & Put carrot in yellow plate & 1.00   & --    & 1.00   & 1.00   & --    & 1.00   & 1.00   & --    & 1.00   & 1.00   & --    & 1.00 \\
          & put blue cube on green cube & 0.90    & --    & 0.80    & 0.90    & --    & 0.90    & 1.00   & --    & 0.90    & 1.00   & --    & 0.90 \\
          & put eggplant in blue bowl & 1.00   & --    & 0.90    & 0.90    & --    & 0.90    & 0.90    & --    & 0.90    & 1.00   & --    & 1.00 \\
          & Put mango in basket & 0.90    & --    & 0.90    & 0.90    & --    & 0.90    & 0.90    & --    & 0.90    & 1.00   & --    & 0.90 \\
    \midrule
    \multicolumn{1}{c}{\multirow{3}[2]{*}{Put \texttt{U} in \texttt{U}}} & Put eggplant in pink bowl & 0.90    & --    & 0.80    & 0.90    & --    & 0.90    & 1.00   & --    & 1.00   & 1.00   & --    & 1.00 \\
          & Put apple in yellow plate & 0.70    & --    & 0.30    & 1.00   & --    & 0.80    & 1.00   & --    & 0.90    & 0.90    & --    & 0.90 \\
          & Put red cube on yellow cube & 0.70    & --    & 0.70    & 0.90    & --    & 0.70    & 0.90    & --    & 0.80    & 0.90    & --    & 0.80 \\
    \midrule
    \multicolumn{1}{c}{\multirow{6}[2]{*}{Put \texttt{U} in \texttt{S}}} & Put strawberry in green plate & 0.00     & 0.90    & 0.00     & 0.70    & 0.00     & 0.70    & 0.70    & 0.10    & 0.70    & 0.90    & 0.10    & 0.90 \\
          & Put grape in green plate & 0.00     & 0.90    & 0.00     & 0.70    & 0.20    & 0.50    & 0.80    & 0.20    & 0.70    & 0.90    & 0.10    & 0.90 \\
          & Put orange in green plate & 0.00     & 1.00   & 0.00     & 0.30    & 0.20    & 0.40    & 0.70    & 0.30    & 0.60    & 1.00   & 0.00     & 1.00 \\
          & Put mango in yellow plate & 0.60    & 0.20    & 0.50    & 0.70    & 0.10    & 0.60    & 0.70    & 0.10    & 0.60    & 0.80    & 0.00     & 0.80 \\
          & Put orange in yellow plate & 0.50    & 0.30    & 0.40    & 0.70    & 0.20    & 0.50    & 0.60    & 0.10    & 0.50    & 1.00   & 0.10    & 1.00 \\
          & Put orange in yellow plate & 0.20    & 0.30    & 0.10    & 0.60    & 0.30    & 0.40    & 0.70    & 0.20    & 0.50    & 1.00   & 0.20    & 1.00 \\
    \midrule
    \multicolumn{1}{c}{\multirow{3}[2]{*}{Move \texttt{U} to \texttt{U}}} & Move strawberry to banana & 0.50    & 0.50    & 0.10    & 0.60    & 0.20    & 0.20    & 0.60    & 0.00     & 0.40    & 1.00   & 0.00     & 1.00 \\
          & Move orange to banana & 0.50    & 0.30    & 0.10    & 0.80    & 0.00     & 0.50    & 0.70    & 0.20    & 0.50    & 0.90    & 0.20    & 0.90 \\
          & Move cube to orange & 0.50    & 0.50    & 0.20    & 0.80    & 0.20    & 0.60    & 0.70    & 0.20    & 0.60    & 1.00   & 0.00     & 1.00 \\
    \midrule
    \multicolumn{2}{c}{\textbf{Average}} & 0.56 & 0.60 & 0.42 & 0.77 & 0.16 & 0.63 & 0.83 & 0.16 & 0.71 & \textbf{0.96} & \textbf{0.08} & \textbf{0.94} \\
    \bottomrule
    \end{tabular}%
    }
    \vspace{0.2cm}
    \begin{tablenotes}
    \footnotesize
    \item \textbf{Note:} PSR indicates partial success rate with grasping the correct object; Dist. indicates the distractor-grasping rate; SR indicates overall task success rate. All rates are decimals in $[0,1]$.
    \end{tablenotes}
  \label{tab:vlaComp}%
\end{table*}
\begin{table}[t]
  \caption{Detailed Distractor Configurations for Unseen Object Manipulation Scenarios}
  \label{tab:distractor_details}
  \centering
  \begin{tabular}{cccc}
    \toprule
    \multirow{2}{*}{Task Format} & \multicolumn{2}{c}{Objects of Interest} & \multirow{2}{*}{Distractor(s)} \\
    \cmidrule(lr){2-3} 
     & Target & Container & \\
    \midrule
    \multirow{6}{*}{Put \texttt{U} in \texttt{S}} & strawberry & green plate & apple \\
     & grape & green plate & eggplant \\
     & orange & green plate & banana \\
     & mango & yellow plate & apple \\
     & orange & yellow plate & apple and grape \\
     & orange & yellow plate & apple, grape and mango \\
    \midrule
    \multirow{3}{*}{Move \texttt{U} to \texttt{U}} & strawberry & banana & apple \\
     & orange & banana & apple \\
     & cube & orange & banana \\
    \bottomrule
  \end{tabular}
\end{table}
We compare our UniManip framework against three state-of-the-art VLA methods: $\pi_0$~\cite{black2024pi_0}, NORA~\cite{hung2025nora1}, and NORA-1.5~\cite{hung2025nora}. These baselines represent the current leading end-to-end approaches in robotic manipulation using vision-language models. 
To facilitate fine-tuning of the baseline models, we curated a dataset comprising 1,000 demonstration episodes collected on a Galaxea A1 robotic platform. While the baselines utilize a dual-camera configuration, consisting of a wrist-mounted Intel RealSense D435 and a third-person RealSense D515, our proposed method still operates using only the single wrist camera.
Data were acquired via isomorphic teleoperation, covering nine distinct pick-and-place tasks with a distribution of approximately 110 episodes per task. To satisfy the input requirements of the VLA baselines, this experimental setup incorporates a third-view camera for global context and a wrist camera for localized manipulation cues. To ensure the robustness of the learned policies, we employed a stochastic setup: objects were placed in unstructured configurations across the workspace without a predefined sequence, requiring the models to generalize across varied initial states and spatial relations.
\begin{figure}
    \centering
    \includegraphics[width=1\linewidth]{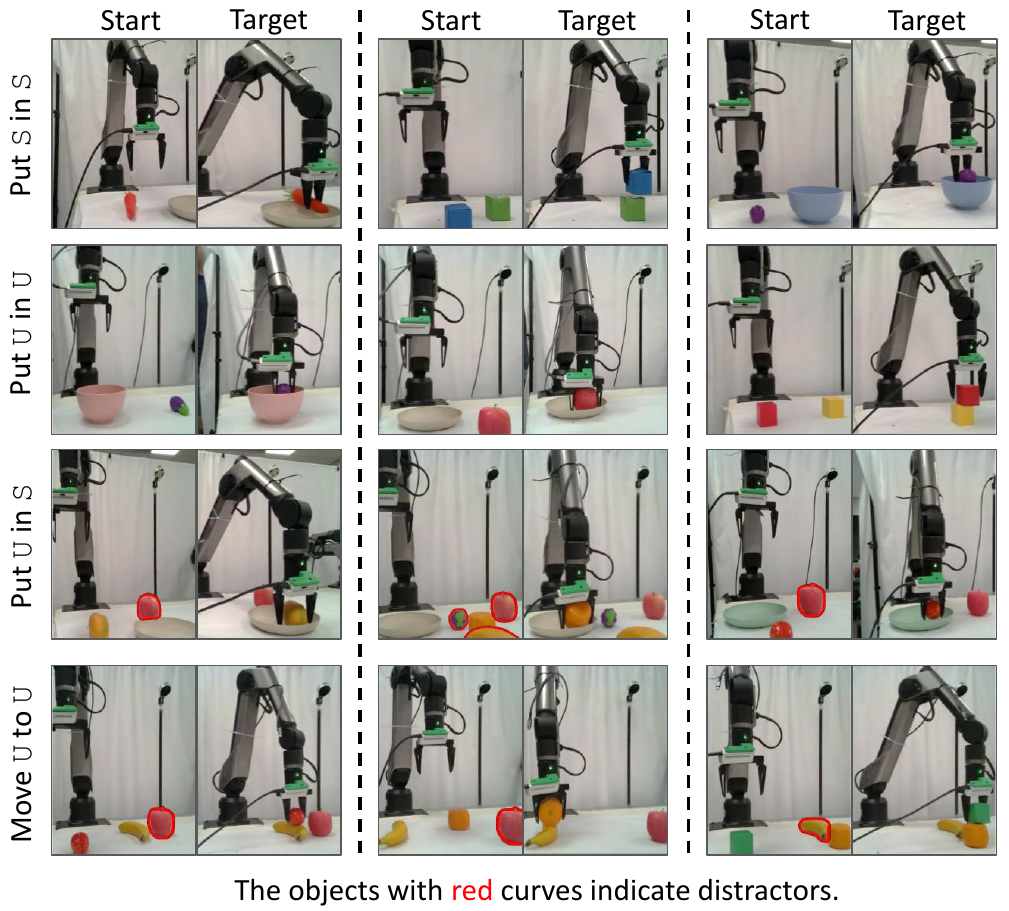}
    \caption{The comparison scenarios with VLA models. The distractors (unseen objects during model training) are marked with red curves.}
    \label{fig:vla_comp_settings}
\end{figure}
The evaluation (shown in Table~\ref{tab:vlaComp}) focuses on a suite of manipulation tasks that vary in complexity and object familiarity. With the notation of \texttt{U} and \texttt{S} representing ``Unseen" and ``Seen" objects respectively, we categorize the tasks into four main types: ``Put \texttt{S} in \texttt{S}," ``Put \texttt{U} in \texttt{U}," ``Put \texttt{U} in \texttt{S}," and ``Move \texttt{U} to \texttt{U}." As illustrated in Fig.~\ref{fig:vla_comp_settings}, seen objects are those that were part of the training dataset, while unseen objects are novel items not encountered during training. To rigorously test the generalization capabilities of each method, we introduce distractor objects in the environment during the unseen tasks, as detailed in Table~\ref{tab:distractor_details}.
Each task category is designed to test the system's ability to generalize to new objects and scenarios. The performance metrics used for comparison include the partial success rate (PSR), the distractor grasping rate (Dist.), and the success rate (SR). These metrics provide a comprehensive assessment of each method's effectiveness in executing the specified manipulation tasks.

The results show that UniManip consistently outperforms end-to-end VLA baselines, with an average success rate (SR) of $0.94$ compared to $0.71$ for NORA-1.5-FAST (the strongest baseline). This gap is most pronounced in the \texttt{U}-involving categories (\texttt{U} in \texttt{S} and \texttt{U} to \texttt{U}), where the policy must generalize beyond the demonstrated object set.

These gains are enabled by two architectural choices.

\textbf{Grounded task representation reduces distractor errors.} UniManip achieves a much lower distractor grasping rate of $0.08$. This follows from the Bi-level AOG: the agentic logic layer queries the SOSG for object-centric constraints (identity, role, and spatial relations) before committing to a manipulation primitive. In contrast, end-to-end VLAs can overfit to co-occurrence patterns in demonstrations and may drift toward salient distractors when the distribution shifts.

\textcolor{lblue}{\textbf{Explicit task-to-motion bridging improves execution reliability.} Even when the target is correctly identified (high PSR), end-to-end policies can fail because they become stuck in local optima or collide during approach. UniManip parameterizes each operation through (a) conservative occupancy completion under single-view occlusion and (b) safety-aware planning, followed by relaxed IK tracking when strict IK is infeasible. This makes the final motion physically consistent with the scene, improving SR beyond purely semantic correctness.}

\subsection{Robustness Evaluation in Cluttered Scenarios}
\label{sec:experiments_hier}
We further evaluate UniManip against representative hierarchical open-vocabulary manipulation frameworks, including ReKep~\cite{huang2025rekep}, MOKA~\cite{liu2024moka}, and VoxPoser~\cite{huang2023voxposer}, as shown in Table~\ref{tab:hierarComp}. The benchmark focuses on cluttered desktop scenes with multiple distractors (Table~\ref{table:distractor_configs}), which stress both semantic grounding (selecting the correct target/container) and geometric feasibility (collision-free reaching and placement under clutter).
The evaluation results report the task success rate (TSR). UniManip achieves $0.83$, substantially higher than ReKep ($0.48$), MOKA ($0.58$), and VoxPoser ($0.55$). We attribute the gain to two factors:

\textcolor{lblue}{Table~\ref{tab:hierarComp} further adds RoboEXP, a graph-based baseline, and a controlled UniManip w/o Graph variant. RoboEXP attains $0.39$ and UniManip w/o Graph attains $0.40$ over the same 80 trials, while full UniManip attains $0.83$. Because the no-graph variant preserves the full low-level stack and recovery budget, the $0.43$ absolute difference from full UniManip isolates the contribution of retaining operation context, verified SOSG state, and conditional graph restructuring rather than attributing the gain solely to perception or motion generation.}

\textbf{The AOG architecture enhances task resilience through plan adaptation.} In clutter, a correct high-level plan is insufficient: intermediate outcomes must be verified, and the remaining plan must be updated when the world changes (e.g., an object is displaced during grasp). UniManip’s AOG explicitly stores operation-level context (tool inputs/outputs and expected post-conditions) and couples it with the objects' states, allowing the system to detect deviations and either locally repair or globally replan. Baselines that rely on a single-pass plan (or weak verification) tend to cascade failures once an early step drifts.

\textbf{Task-to-motion bridge reduces geometric brittleness.} Current hierarchical methods produce goal representations (keypoints, value maps, or constraints) but can still be brittle when the generated motion is marginally infeasible or unsafe. UniManip explicitly accounts for occlusion-induced uncertainty through conservative occupancy completion and plans with safety-aware clearance constraints, then uses relaxed IK to avoid dead-ends near singularities. This reduces collisions and IK failures that otherwise dominate in clutter.

Finally, UniManip’s compact tool interface (as a demonstration in Fig.~\ref{fig:spatialPipeline}) reduces the burden on the foundation model: the VLM is primarily used for high-level decomposition and verification, while geometric feasibility is handled by dedicated planning modules. This makes the pipeline compatible with smaller VLMs (e.g., Qwen-3-VL-4B~\cite{yang2025qwen3} utilized in this paper) and supports practical deployment with limited onboard compute.

\begin{table}[t]
  \centering
  \caption{\textcolor{lblue}{Comparison with Hierarchical and Graph-Based Methods on Cluttered Desktop Tasks. Each entry is the success rate as a decimal over 10 randomized trials.}}
  \label{tab:hierarComp}
  \setlength{\tabcolsep}{2.5pt} %
  \resizebox{\columnwidth}{!}{%
    \begin{tabular}{lcccccc}
      \toprule
      Task & ReKep~\cite{huang2025rekep} & MOKA~\cite{liu2024moka} & VoxPoser~\cite{huang2023voxposer} & \textcolor{lblue}{RoboEXP~\cite{roboexp2024}} & \textcolor{lblue}{\shortstack{Ours\\w/o Graph}} & \textbf{Ours} \\
      \midrule
      \textcolor{lblue}{PW-OB} & 0.60 & 0.60 & 0.50 & \textcolor{lblue}{0.40} & \textcolor{lblue}{0.20} & \textbf{0.80} \\
      \textcolor{lblue}{PW-AP} & 0.60 & 0.60 & 0.70 & \textcolor{lblue}{0.60} & \textcolor{lblue}{0.70} & \textbf{0.90} \\
      \textcolor{lblue}{PW-TB} & 0.20 & 0.80 & 0.70 & \textcolor{lblue}{0.40} & \textcolor{lblue}{0.00} & \textbf{0.90} \\
      \textcolor{lblue}{PW-ST} & 0.40 & 0.20 & \textbf{0.60} & \textcolor{lblue}{0.00} & \textcolor{lblue}{0.20} & \textbf{0.60} \\
      \textcolor{lblue}{PW-PP} & 0.40 & 0.60 & 0.40 & \textcolor{lblue}{0.50} & \textcolor{lblue}{0.40} & \textbf{0.70} \\
      \textcolor{lblue}{PW-CR} & 0.80 & 0.60 & 0.60 & \textcolor{lblue}{0.50} & \textcolor{lblue}{0.40} & \textbf{0.90} \\
      \textcolor{lblue}{PW-BN} & 0.40 & 0.60 & 0.40 & \textcolor{lblue}{0.50} & \textcolor{lblue}{0.50} & \textbf{0.90} \\
      \textcolor{lblue}{PW-CU} & 0.40 & 0.60 & 0.50 & \textcolor{lblue}{0.20} & \textcolor{lblue}{0.80} & \textbf{0.90} \\
      \midrule
      \textbf{Average}         & 0.48 & 0.58 & 0.55 & \textcolor{lblue}{0.39} & \textcolor{lblue}{0.40} & \textbf{0.83} \\
      \bottomrule
    \end{tabular}%
  }
  \vskip 3pt
  \begin{tablenotes}[flushleft]\footnotesize
    \item[] \textbf{Note:} PW denotes ``put''; the suffixes identify the target/container pair: OB (orange--blue bowl), AP (apple--beige plate), TB (tomato--basket), ST (strawberry--basket), PP (pepper--blue plate), CR (carrot--green plate), BN (banana--blue plate), and CU (cucumber--green plate).
  \end{tablenotes}
\end{table}
\begin{table}[t]
\centering
\caption{Detailed Distractor Configurations for Cluttered Object Manipulation Scenarios}
\label{table:distractor_configs}
\resizebox {\columnwidth}{!}{%
\begin{tabular}{llc}
\toprule
\multicolumn{2}{c}{Objects of Interest} & \multirow{2}{*}{Distractor(s)} \\ \cline{1-2}
Target & Container & \\ 
\midrule
orange & blue bowl & strawberry, apple, tomato, green plate, azure plate \\
apple & beige plate & orange, strawberry, tomato, blue bowl, green plate \\
tomato & basket & carambola, orange, blue plate, blue bowl \\
strawberry & basket & carambola, orange, azure plate, blue plate, blue bowl \\
pepper & blue plate & carrot, azure plate, blue bowl, green plate \\
carrot & green plate & eggplant, azure plate, blue bowl, blue plate \\
banana & blue plate & eggplant, azure plate, blue bowl, green plate \\
cucumber & green plate & banana, blue plate \\
\bottomrule
\end{tabular}
}
\end{table}

\begin{table}[t]
  \centering
  \caption{\textcolor{lblue}{LIBERO-Pro comparison. Success rates are reported as decimals for position (Pos.) and task (Task) perturbations in the Object, Goal, Spatial, and Long suites. These metrics respectively test geometric pose generalization and task-level reasoning.}}
  \resizebox{\columnwidth}{!}{%
  \begin{tabular}{lcccccccc}
  \toprule
  & \multicolumn{2}{c}{Object} & \multicolumn{2}{c}{Goal} & \multicolumn{2}{c}{Spatial} & \multicolumn{2}{c}{Long} \\
  \textcolor{lblue}{Method} & \textcolor{lblue}{Pos.} & \textcolor{lblue}{Task} & \textcolor{lblue}{Pos.} & \textcolor{lblue}{Task} & \textcolor{lblue}{Pos.} & \textcolor{lblue}{Task} & \textcolor{lblue}{Pos.} & \textcolor{lblue}{Task} \\
  \midrule
  \textcolor{lblue}{MolmoAct~\cite{molmoact2025}} & \textcolor{lblue}{0.06} & \textcolor{lblue}{0.00} & \textcolor{lblue}{0.00} & \textcolor{lblue}{0.00} & \textcolor{lblue}{0.00} & \textcolor{lblue}{0.00} & \textcolor{lblue}{0.00} & \textcolor{lblue}{0.06} \\
  \textcolor{lblue}{NORA~\cite{hung2025nora1}} & \textcolor{lblue}{0.00} & \textcolor{lblue}{0.00} & \textcolor{lblue}{0.00} & \textcolor{lblue}{0.00} & \textcolor{lblue}{0.00} & \textcolor{lblue}{0.00} & \textcolor{lblue}{0.00} & \textcolor{lblue}{0.00} \\
  \textcolor{lblue}{X-VLA~\cite{xvla2025}} & \textcolor{lblue}{0.02} & \textcolor{lblue}{0.08} & \textcolor{lblue}{0.01} & \textcolor{lblue}{0.09} & \textcolor{lblue}{0.00} & \textcolor{lblue}{0.00} & \textcolor{lblue}{0.00} & \textcolor{lblue}{0.10} \\
  \textcolor{lblue}{OpenVLA~\cite{kim2025openvla}} & \textcolor{lblue}{0.00} & \textcolor{lblue}{0.00} & \textcolor{lblue}{0.00} & \textcolor{lblue}{0.00} & \textcolor{lblue}{0.00} & \textcolor{lblue}{0.00} & \textcolor{lblue}{0.00} & \textcolor{lblue}{0.00} \\
  \textcolor{lblue}{$\pi_0$~\cite{black2024pi_0}} & \textcolor{lblue}{0.00} & \textcolor{lblue}{0.00} & \textcolor{lblue}{0.00} & \textcolor{lblue}{0.00} & \textcolor{lblue}{0.00} & \textcolor{lblue}{0.00} & \textcolor{lblue}{0.00} & \textcolor{lblue}{0.00} \\
  \textcolor{lblue}{$\pi_{0.5}$~\cite{black2025pi_}} & \textcolor{lblue}{0.17} & \textcolor{lblue}{0.01} & \textcolor{lblue}{\textbf{0.38}} & \textcolor{lblue}{0.00} & \textcolor{lblue}{0.20} & \textcolor{lblue}{0.01} & \textcolor{lblue}{0.08} & \textcolor{lblue}{0.01} \\
  \textcolor{lblue}{CaP-Agent0~\cite{fu2026cap}} & \textcolor{lblue}{0.22} & \textcolor{lblue}{0.18} & \textcolor{lblue}{0.26} & \textcolor{lblue}{0.17} & \textcolor{lblue}{0.12} & \textcolor{lblue}{0.14} & \textcolor{lblue}{--} & \textcolor{lblue}{--} \\
  \textcolor{lblue}{UniManip} & \textcolor{lblue}{\textbf{0.77}} & \textcolor{lblue}{\textbf{0.85}} & \textcolor{lblue}{0.33} & \textcolor{lblue}{\textbf{0.29}} & \textcolor{lblue}{\textbf{0.32}} & \textcolor{lblue}{\textbf{0.37}} & \textcolor{lblue}{\textbf{0.21}} & \textcolor{lblue}{\textbf{0.17}} \\
  \bottomrule
  \end{tabular}}
  \label{tab:libero_pro}
\end{table}

\textcolor{lblue}{Following the standard LIBERO-Pro benchmark~\cite{zhou2025libero}, Table~\ref{tab:libero_pro} focuses the comparison on the perturbations most diagnostic of physically grounded manipulation: \textit{Pos.} assesses geometric pose generalization, while \textit{Task} evaluates task-level reasoning. UniManip achieves the highest performance on Object, Spatial, and Long Pos variations, as well as on Task success across all four evaluation suites. The only exception is Goal Pos, where UniManip achieves 0.33 compared to 0.38 by $\pi_{0.5}$; this margin stems from the baseline model’s extensive pre-training, which specifically bolsters its robustness against position variations. Overall, these benchmark results demonstrate the efficacy of our physically grounded geometric reasoning and AOG-based task execution under standardized task and action interfaces.}

\begin{figure}
    \centering
    \includegraphics[width=\linewidth]{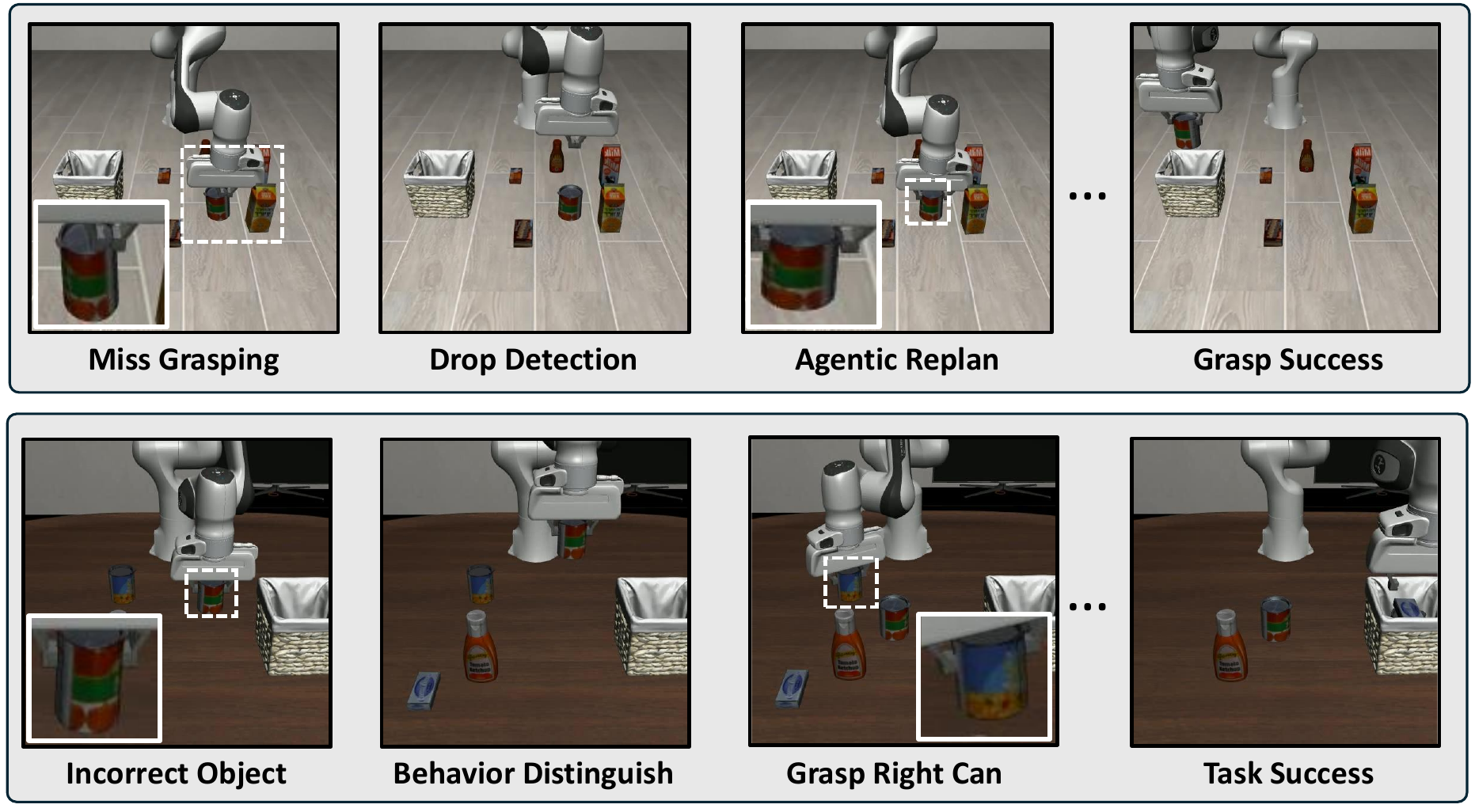}
    \caption{\textcolor{lblue}{Qualitative case studies of UniManip's AOG for closed-loop failure recovery and target verification. Top: UniManip detects the object slipping during lifting, autonomously triggers agentic replanning, and successfully executes a regrasp. Bottom: When an incorrect distractor is initially engaged, the agent verifies the object attributes against the task instruction, rectifies the target selection, and completes the placement task.}}
    \label{fig:case_study}
\end{figure}

\textcolor{lblue}{Specifically, the utility of the AOG is evaluated in the empirical study, to show the benefit of the graph-based task representation and reasoning. The AOG allows for explicit verification of intermediate states and enables local repair or global replanning when deviations occur. As illustrated in Fig.~\ref{fig:case_study}, the AOG-based approach detects improper grasps and initiates a recovery routine, whereas baseline methods fail to recognize the error and proceed, resulting in task failure. Furthermore, if the target object is misidentified, the AOG verifies its identity and executes corrective actions to ensure proper manipulation.}

\begin{figure*}[t]
    \centering
    \includegraphics[width=0.95\textwidth]{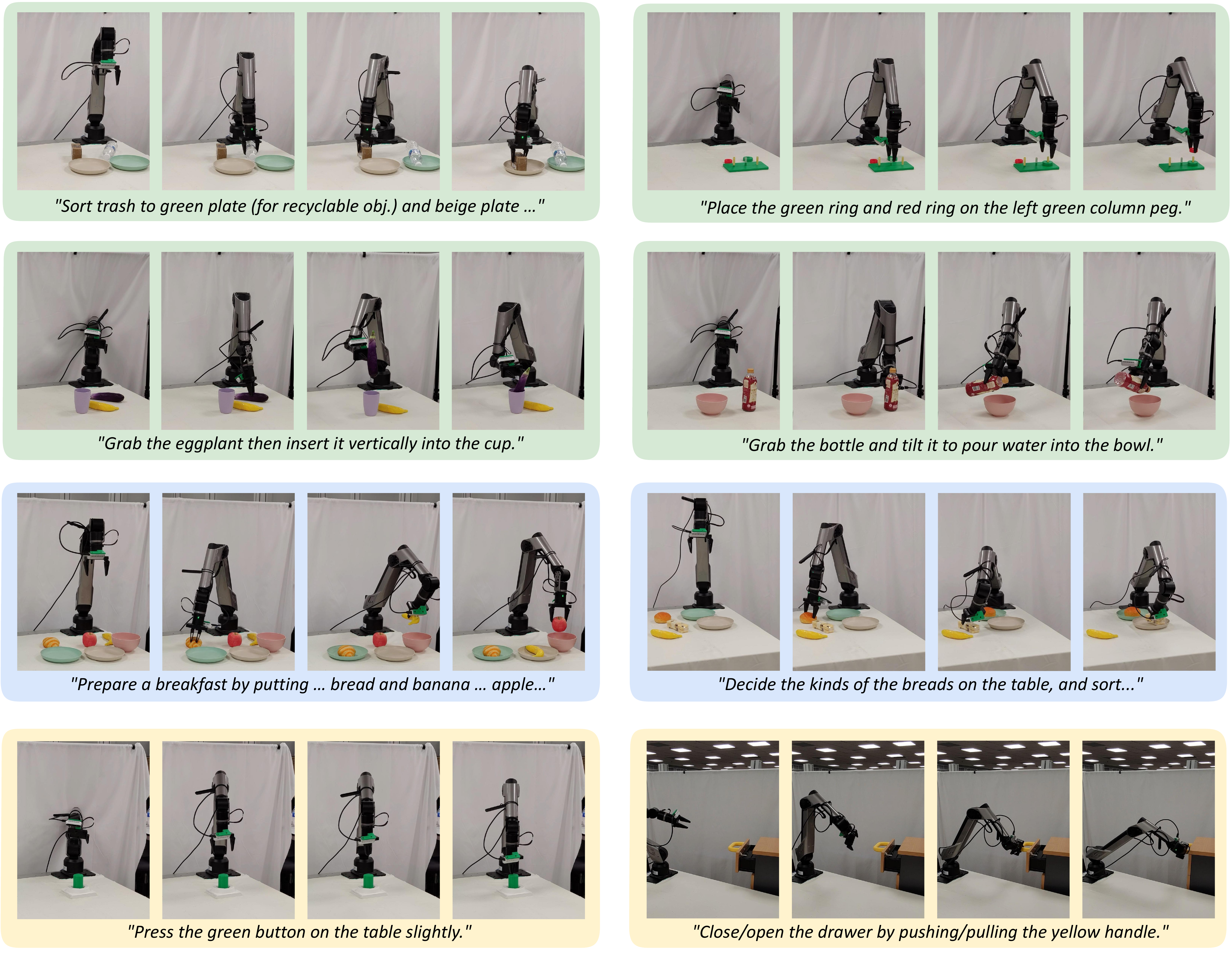} 
    \caption{Demonstration of UniManip executing various manipulation tasks in a cluttered desktop environment. The robot successfully performs tasks such as sorting trash, inserting objects, pouring water, preparing breakfast, and opening/closing drawers, showcasing its versatility and robustness.}
    \label{fig:long_horizon_dexterous_tasks}
\end{figure*}
\begin{table}[t]
  \centering
  \caption{\textbf{Ablation Settings.} We evaluate three configurations (V1, V2, V3) to analyze the contribution of each module. (\ding{51}: Included, \ding{55}: Excluded)}
  \begin{tabularx}{\linewidth}{X ccc}
    \toprule
    \textbf{Feature} & \textbf{V1} & \textbf{V2} & \textbf{V3} \\
    \midrule
    Parameter-relaxed Inverse Kinematics (RIK) & \ding{51} & \ding{51} & \ding{51} \\
    Collision-Free Planning (CFP)               & \ding{55} & \ding{51} & \ding{51} \\
    Failure Detection \& Recovery (FDR)        & \ding{55} & \ding{55} & \ding{51} \\
    \bottomrule
  \end{tabularx}
  \label{tab:ablation_settings}
\end{table}
\subsection{Evaluation on Long-Horizon and Dexterous Tasks}
\label{sec:experiments_longhorizon}
While the previous comparisons mainly test short-horizon pick-and-place capability, we additionally evaluate UniManip on long-horizon and interaction-rich tasks that are typically out-of-scope for VLA and the hierarchical baselines. These scenarios require (1) sequential composition of heterogeneous skills, (2) precise contact control and geometric safety in clutter, and (3) closed-loop recovery when intermediate steps fail. The execution traces for these representative manipulation tasks are visualized in Fig.~\ref{fig:long_horizon_dexterous_tasks}.

To evaluate the versatility of UniManip, we conducted experiments across eight distinct scenarios categorized by their physical properties and interaction requirements. \textbf{Rigid object manipulation} tasks focus on precision and geometry-aware placement, including the sorting of waste items (e.g., plastic bottles and batteries) into designated bins, the insertion of geometric shapes into poled bases, and the placement of elongated objects into narrow apertures. This category also evaluates controlled interactions, such as pouring water from a bottle into a bowl. We further assess \textbf{deformable object interaction} through the sorting of diverse bread types and a multi-stage breakfast preparation sequence requiring the handling of soft food items through coordinated pick-and-place and tool-use motions. Finally, \textbf{articulated-object interaction} is evaluated through button-pressing under tight positional tolerances and the execution of drawer opening and closing sequences involving robust state verification.

To isolate which components drive performance, we ablate three execution-critical modules: Parameter-relaxed Inverse Kinematics (RIK), Collision-Free Planning (CFP), and Failure Detection \& Recovery (FDR). 
The specific configurations for these variants are detailed in Table~\ref{tab:ablation_settings}, with their respective contributions to overall task success quantified in Table~\ref{tab:ablation_results}.
The ablation results reveal clear causal links between modules and task families.

\textbf{RIK primarily improves reachability robustness.} RIK mitigates \textit{no-solution} IK events near singularities or workspace boundaries by relaxing pose tolerances only when strict IK fails. 
This flexibility is particularly critical for collision-free trajectory tracking (as an example in Fig.~\ref{fig:occupancy_map}) and the manipulation of articulated objects. Even in relatively sparse environments, target poses can be geometrically constrained by the volumetric occupancy of obstacles or the kinematic limits of the assembly, yet marginal deviations often remain functionally admissible for maintaining task progress.

\textcolor{lblue}{\textbf{CFP trades a small amount of completion rate for safety in clutter.} Enabling Collision-Free Planning (V2) reduces the average success rate by $0.01$, consistent with the conservatism of the ESDF planner: it can reject a marginally reachable motion or time out while replanning around a cluttered region. This small aggregate cost is accompanied by a substantial reduction in edge collisions, from $38.9\%$ without CFP to $5.6\%$ with CFP. The improvement is especially evident for the bowl with a deep bottom shown in Fig.~\ref{fig:raw_scenario_grid}, where the planner avoids trajectories that skim or intersect the container edge. Thus, V2 is the safety-constrained operating point used by the full system, accepting a slight completion-rate decrease to reduce unintended contact in clutter.}

\textcolor{lblue}{\textbf{FDR primarily improves long-horizon success by preventing error accumulation.} When a step fails (e.g., a grasp slips or an object is displaced), the AOG-driven verifier updates the SOSG and triggers local repair or replanning. To balance recovery-time overhead and task efficiency, we set the maximum trial number to $R_{\text{max}}=3$. Because most recovery behaviors are local repairs, such as regrasping the object, the average execution time increases by only $14.4\%$ relative to V2 without FDR. These bounded retries convert would-be terminal failures into recoverable states, explaining the gain from V2 to V3 in multi-step tasks such as trash sorting and breakfast preparation.}

Overall, these results show that strong long-horizon performance arises from combining grounded state representations and verification-driven recovery with reliable, safety-aware motion generation, rather than relying solely on open-loop action execution.

\begin{table*}[htbp]
  \centering
  \caption{\textbf{Ablation Results.} Success rates are reported as decimals across different manipulation categories. The configurations V1, V2, and V3 are defined in Table~\ref{tab:ablation_settings}.}
  \renewcommand{\arraystretch}{1.0}
  \begin{tabularx}{\textwidth}{l X X ccc}
    \toprule
    \multirow{2}{*}{\textbf{Category}} & \multirow{2}{*}{\textbf{Task}} & \multirow{2}{*}{\textbf{Objects}} & \multicolumn{3}{c}{\textbf{Variants of UniManip}} \\
    \cmidrule(lr){4-6}
     & & & \textbf{V1} & \textbf{V2} & \textbf{V3} \\
    \midrule
    
    \multirow{4}{*}{\shortstack[l]{Rigid obj.\\ manipulation}} 
          & Sort trash to different container & Plastic bottle, batteries & 0.50 & 0.40 & 0.90 \\
          & Insert bricks & Green ring, cube, base with poles & 0.40 & 0.45 & 0.70 \\
          & Insert rod-shaped object & Eggplant, banana & 0.50 & 0.40 & 0.80 \\
          & Pour water & Bottle, bowl & 0.20 & 0.10 & 0.60 \\
    \midrule
    
    \multirow{2}{*}{\shortstack[l]{Deformable obj.\\ manipulation}} 
          & Prepare breakfast & Bread, apple, banana, plate, bowl & 0.60 & 0.40 & 0.90 \\
          & Sort breads into different plates & Bread roll, loaf of bread, coconut bread & 0.40 & 0.50 & 0.70 \\
    \midrule
    
    \multirow{3}{*}{\shortstack[l]{Articulated obj.\\ manipulation}} 
          & Press a button & Green cylinder & 0.90 & 0.90 & 1.00 \\
          & Open the drawer & Cabinet with opened drawer & 0.60 & 0.80 & 0.80 \\
          & Close the drawer & Cabinet with closed drawer & 0.60 & 0.60 & 0.70 \\
    \midrule
    
    \multicolumn{3}{l}{\textbf{Average Success Rate across the 90 Trials}} & \textbf{0.52} & \textbf{0.51} & \textbf{0.80} \\
    \bottomrule
  \end{tabularx}
  \label{tab:ablation_results}
\end{table*}
\subsection{\textcolor{lblue}{Cross-Embodiment Generalization of UniManip}}
\label{subsec:cross_embodiment}
\label{subsec:experiments_mobile}
\textcolor{lblue}{We evaluate transfer from the fixed-base Galaxea A1 with a parallel-jaw gripper along three embodiment axes: arm kinematics, end-effector morphology, and base mobility. In each case, the AOG task representation, operation sequencing, verification, and recovery interfaces are retained, while the hardware-facing module required by the new embodiment is adapted. Table~\ref{tab:cross_embodiment} reports the fixed-base UR7e trials, and Table~\ref{tab:mobile_manipulation} reports the mobile-platform trials.}

\subsubsection{\textcolor{lblue}{Kinematic Arm Transfer: Porting to the UR7e Manipulator}}
\textcolor{lblue}{We replace the Galaxea A1 arm with a fixed-base, 6-DoF UR7e equipped with a Robotiq RF-85 parallel-jaw gripper. UniManip separates task-level operation selection from joint-space execution: the planner continues to provide collision-free end-effector waypoints, while RIK, the task-to-motion bridge in Fig.~\ref{fig:overview}, maps those waypoints to the new arm. At the kinematic layer, porting consists of substituting the UR7e kinematic tree and joint limits from its standard URDF into the RIK solver. The tool-frame transform is calibrated and the robot command interface is mapped to the UR7e controller for physical execution; neither change requires retraining or redesigning the core agent pipeline.}

\textcolor{lblue}{The UR7e--Robotiq configuration performs two interaction-rich tasks (Table~\ref{tab:cross_embodiment}): tilting a bottle to pour water and preparing a meal by placing a banana on a plate and an apple in a bowl. Across 20 trials per task, success rates are 0.65 and 0.80, respectively. These trials test whether the same task decomposition and waypoint-synthesis interfaces remain executable under a different arm workspace and joint configuration.}

\begin{table}[t]
  \centering
  \caption{\textcolor{lblue}{Cross-embodiment hardware evaluation. Success rates are decimals; times are mean $\pm$ sample standard deviation over all recorded trials.}}
  \resizebox{\columnwidth}{!}{%
  \begin{tabular}{llccc}
  \toprule
  \textcolor{lblue}{Embodiment} & \textcolor{lblue}{Task} & \textcolor{lblue}{Success} & \textcolor{lblue}{Total time (s)} & \textcolor{lblue}{VLM time (s)} \\
  \midrule
  \textcolor{lblue}{UR7e + RF-85} & \textcolor{lblue}{Pour water} & \textcolor{lblue}{0.65} & \textcolor{lblue}{41.22 $\pm$ 14.14} & \textcolor{lblue}{5.49 $\pm$ 1.15} \\
  \textcolor{lblue}{UR7e + RF-85} & \textcolor{lblue}{Prepare meal} & \textcolor{lblue}{0.80} & \textcolor{lblue}{53.52 $\pm$ 17.98} & \textcolor{lblue}{7.87 $\pm$ 2.06} \\
  \textcolor{lblue}{UR7e + Inspire} & \textcolor{lblue}{Block to drawer} & \textcolor{lblue}{0.53} & \textcolor{lblue}{109.48 $\pm$ 25.26} & \textcolor{lblue}{6.90 $\pm$ 0.61} \\
  \bottomrule
  \end{tabular}}
  \label{tab:cross_embodiment}
\end{table}

\subsubsection{\textcolor{lblue}{End-Effector Morphology Adaptation: Dexterous Hand Manipulation}}
\begin{figure}
    \centering
    \includegraphics[width=0.95\linewidth]{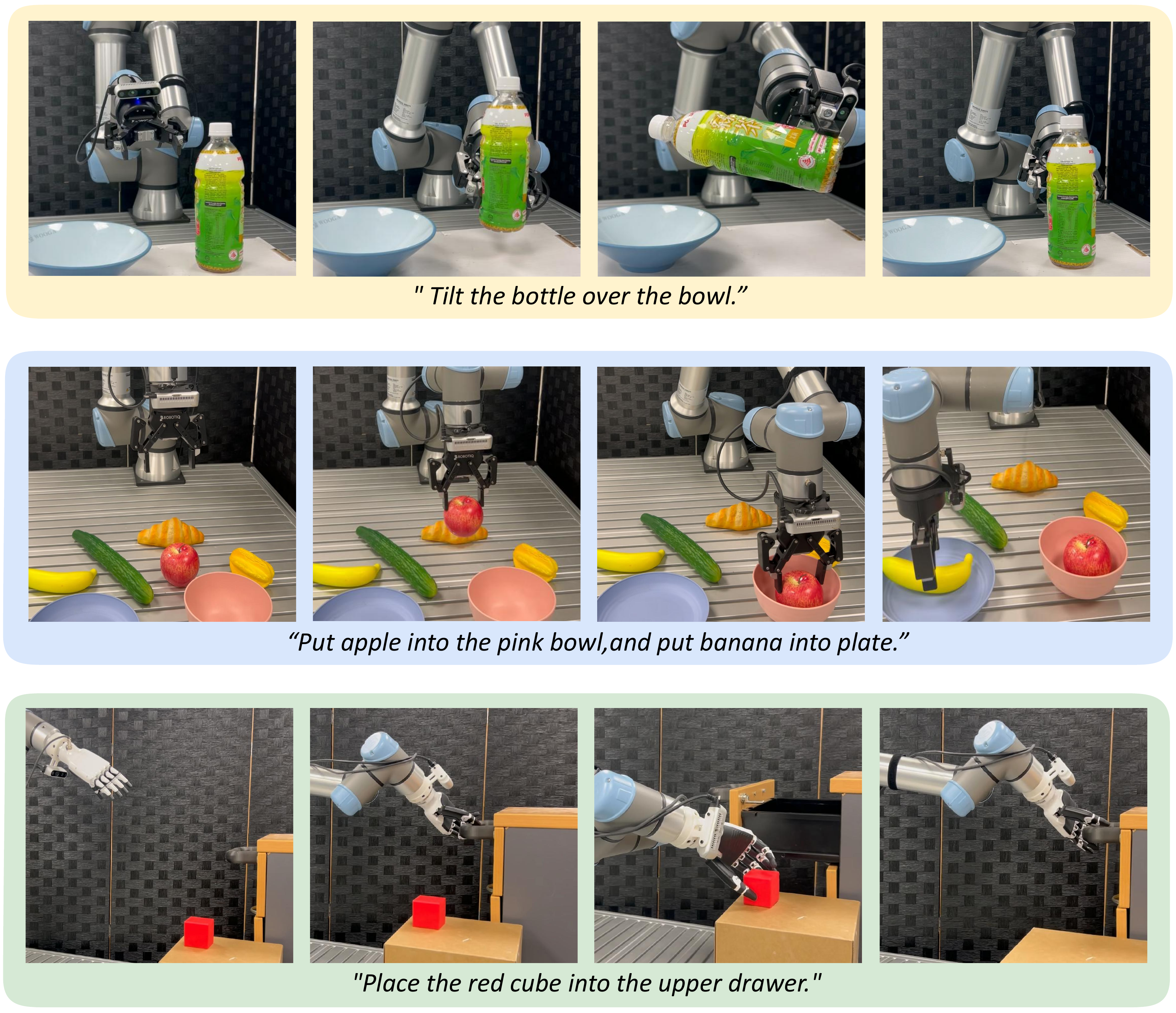}
    \caption{\textcolor{lblue}{Cross-embodiment evaluation on a UR7e arm with a gripper and an Inspire multi-fingered hand, respectively. The tasks need tilting, dexterous grasping, pulling, pushing, and precise placement. The same AOG is used across embodiments and tasks.}}
    \label{fig:cross_embodiment}
\end{figure}

\textcolor{lblue}{To test end-effector transfer, we replace the UR7e's parallel-jaw gripper with an Inspire multi-fingered hand (Fig.~\ref{fig:cross_embodiment}). At the modular grasp-pose perception interface, the algorithmic change is to exchange the standard 6-DoF parallel-grasp generator for a multi-fingered affordance synthesizer, such as AnyDexGrasp~\cite{fang2025anydexgrasp}, while preserving the downstream grasp-pose interface. The hand-specific action adapter maps grasp and release commands to finger actuation, and the hand's tool frame is calibrated. The AOG's task sequence, SOSG state updates, collision-free waypoint synthesis, and verification logic use the same interfaces as for the parallel-jaw embodiment, without task-level retraining.}

\textcolor{lblue}{The Inspire-hand evaluation comprises opening a drawer and placing a block inside it, a composite task requiring articulated-object interaction followed by precise placement. It attains 0.53 success over 15 recorded trials, with mean execution time $109.48\pm25.26$~s (Table~\ref{tab:cross_embodiment}). The reported hand trial specifically tests the change in grasp morphology.}

\subsubsection{\textcolor{lblue}{Mobility Expansion: Whole-Body Manipulation in Office Environments}}
\begin{figure*}
    \centering
    \includegraphics[width=0.95\linewidth]{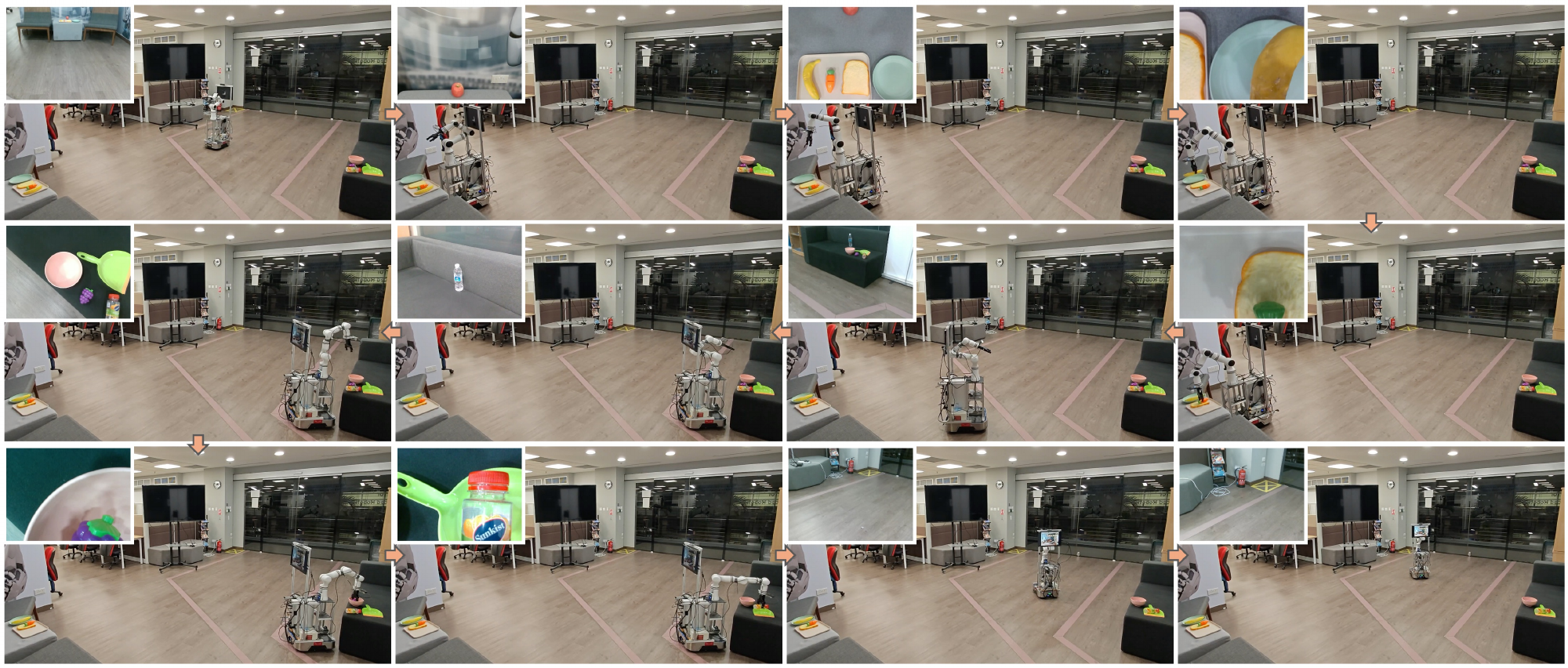}
    \caption{\textcolor{lblue}{Mobile-manipulation demonstration with a Realman RM65 arm on a mobile base. The AOG instantiates navigation and manipulation operations, while SOSG state updates close the loop across docking points in an office environment.}}
    \label{fig:mobile_manip_demo}
\end{figure*}

\textcolor{lblue}{We mount a Realman RM65 arm on a differential-drive mobile base (Fig.~\ref{fig:robot_hardware}) and evaluate office meal-preparation and workspace-cleanup tasks from randomized docking configurations (Fig.~\ref{fig:mobile_manip_demo}). Whole-body coordination is introduced by adding the navigation primitive $\mathbf{N}(\mathcal{O}_{target},\hat d_{obj},d_{goal})$ to the available tool library $\mathcal{T}_c$ for $c=\mathtt{mobile}$. The AOG selects a target entity $\mathcal{O}_{target}$ from the SOSG and specifies a desired docking clearance $d_{goal}$; after docking, it invokes the same arm-manipulation tools and verification steps used on the fixed-base platforms.}

\textcolor{lblue}{The mapless navigation primitive uses visual servoing rather than a reconstructed global map. YOLOE~\cite{wang2025yoloe} detects the target in the RGB stream from the eye-in-hand Intel RealSense D435. Registered D435 depth samples within the detected region provide an online estimate $\hat d_{obj}$ of target distance and a camera-frame target centroid. An angular controller regulates yaw velocity $\omega_z$ to align the target with the camera's principal axis, while a linear controller regulates forward velocity $v_x$ to approach the specified clearance. The docking-phase control law is}
{\color{lblue}
\begin{equation}
\begin{cases}
v_x = K_p (z_{c,\mathrm{obj}}-d_{goal}), \\
\omega_z = K_{\theta}\operatorname{atan2}(x_{c,\mathrm{obj}},z_{c,\mathrm{obj}}),
\end{cases}
\end{equation}
}
\textcolor{lblue}{where $z_{c,\mathrm{obj}}$ and $x_{c,\mathrm{obj}}$ denote the target's depth and lateral offset in the camera frame. We use proportional gains $K_p=0.6$ and $K_{\theta}=0.2$. This closed-loop docking step expands the arm's effective workspace by repositioning the base before manipulation.}

\textcolor{lblue}{The mobile evaluation includes 20 trials per target object across randomized office docking configurations, yielding 80 object episodes (Table~\ref{tab:mobile_manipulation}). The mean grasp, conditional placement, and end-to-end task success rates are 0.85, 0.93, and 0.79, respectively, with mean execution time $29.37\pm6.77$~s. The higher placement rate is conditional on successful grasps; grasp acquisition remains the main stage-wise bottleneck. The bottle cleanup task has the lowest task success (0.75), consistent with its constrained approach geometry and the need for repeated visual alignment and arm adjustment. Variation in execution time is consistent with online re-grounding, closed-loop replanning, and base repositioning. Docking is instantiated from the target SOSG node, and grasp execution proceeds only after a reachable observation.}

\begin{table}[t]
  \centering
  \color{lblue}
  \caption{\textcolor{lblue}{Fine-grained mobile-manipulation results on the Realman RM65 platform. Grasp SR and Task SR are calculated over 20 trials per target object; Place SR is conditioned on successful grasps. Execution times are reported as mean $\pm$ standard deviation.}}
  \label{tab:mobile_manipulation}
  \footnotesize
  \setlength{\tabcolsep}{5.5pt}
  \renewcommand{\arraystretch}{1.15}
  \begin{tabular*}{\columnwidth}{@{\extracolsep{\fill}}lcccc}
    \toprule
    \textbf{Target Object} & \textbf{Grasp SR} & \textbf{Place SR} & \textbf{Task SR} & \textbf{Exec. Time (s)} \\
    \midrule
    \multicolumn{5}{l}{\textit{Meal Preparation}} \\
    \quad Banana & 0.80 & 1.00 & 0.80 & $35.05 \pm 6.25$ \\
    \quad Carrot & 0.80 & 1.00 & 0.80 & $24.91 \pm 2.34$ \\
    \midrule
    \multicolumn{5}{l}{\textit{Workspace Cleanup}} \\
    \quad Grape  & 0.90 & 0.89 & 0.80 & $32.53 \pm 5.81$ \\
    \quad Bottle & 0.90 & 0.83 & 0.75 & $24.99 \pm 5.14$ \\
    \midrule
    \textbf{Overall / Mean} & \textbf{0.85} & \textbf{0.93} & \textbf{0.79} & $\mathbf{29.37 \pm 6.77}$ \\
    \bottomrule
  \end{tabular*}
\end{table}

\textcolor{lblue}{This evaluation transfers the manipulation stack without task-specific training or fine-tuning. The mobile platform nevertheless requires the embodiment-specific navigation primitive and tuned docking gains. Base alignment affects arm reachability, and changing docking viewpoints changes the wrist-camera observation. Thus, the results support portability of the AOG reasoning and geometric-manipulation interfaces when the required embodiment adapters are supplied.}
\begin{figure}
    \centering
    \includegraphics[width=\linewidth]{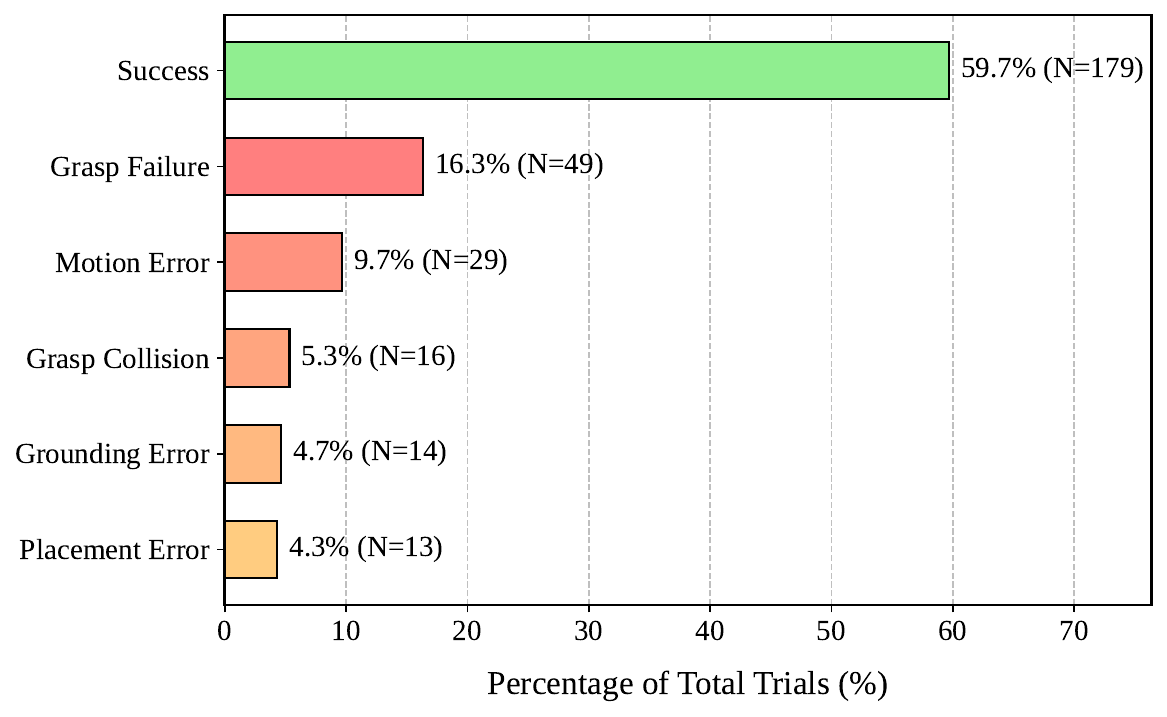}
\caption{\textcolor{lblue}{Breakdown of failure modes observed across all evaluation trials. Physical interaction failures and kinematic constraints are the primary bottlenecks, while semantic grounding remains relatively robust.}}
\label{fig:failure_analysis}
\vspace{-0.3cm}
\end{figure}

\subsection{Failure Analysis}
\label{sec:failure_analysis}
To understand the remaining limitations of our proposed framework, we categorize the failure modes observed across evaluation trials (Fig.~\ref{fig:failure_analysis}), noting that these instances typically stem from operational errors in the execution flow rather than deficiencies in the system’s underlying performance limits.
The system achieved a 0.60 overall success rate on the most challenging long-horizon and dexterous tasks, while the remaining 0.40 failures are grouped into categories before any recovery actions are applied. This isolates intrinsic bottlenecks of the primary perception--planning--execution pipeline, and clarifies where the AOG-based reflection mechanism provides the most leverage.

\textbf{Physical Interaction Failures (Grasp Failure: 0.16, Placement Error: 0.04):}
The most predominant failure mode is \textit{Grasp Failure}, accounting for 0.16 of all trials. These failures primarily stem from the limitations of single-view depth estimation. When dealing with transparent or reflective objects (e.g., plastic bottles), the depth sensor occasionally generates noisy point clouds, leading to inaccurate gripper pose estimation relative to the object's center of mass. Similarly, \textit{Placement Errors} (0.04) often occurred during the release phase, where unstable contact dynamics caused objects to topple upon placement. These physical interaction issues highlight the inherent difficulty of open-loop execution in unmodeled dynamics.

\textbf{Kinematic and Planning Deviations (Motion Error: 0.10, Grasp Collision: 0.05):}
\textit{Motion Errors} accounted for 0.10 of failures. In the context of mobile manipulation, these were largely caused by the coupling errors between the mobile base and the manipulator. If the base navigation stopped slightly outside the optimal workspace, the arm occasionally encountered kinematic singularities or failed to find a valid IK solution for the requested atomic operation. \textit{Grasp Collisions} (0.05) typically occurred in dense environments where the gripper fingers made unintended contact with adjacent obstacles during the approach phase, suggesting a need for more conservative safety margins in the low-level trajectory planner.

\textcolor{lblue}{\textbf{Semantic Grounding Errors (0.05):}
Grounding errors constitute 0.05 of the analyzed failures. This indicates that the structured state representation is useful for the evaluated free-form commands, but does not establish universal semantic robustness. The remaining cases occur when visually similar objects are close together or a spatial description is ambiguous (e.g., distinguishing a \textit{cup} from a \textit{mug}).}

\textcolor{lblue}{\textbf{Role of Reflection:}
The failure distribution motivates verification-driven recovery: grasp, motion, and placement errors are execution mismatches rather than only semantic mistakes. The reflection module can update the observed state, adjust a re-grasp or docking parameter, or replan the remaining operation sequence. It cannot correct a persistent perception error, unavailable grasp, or kinematically unreachable target; these residual failure modes delimit the current system.}

\section{Conclusion}
\textcolor{lblue}{UniManip is a physically grounded agentic operational-graph framework for manipulation within the evaluated task distributions. Its AOG unifies perception, episodic and procedural memory, VLM/LLM-guided high-level planning, and low-level closed-loop execution. The SOSG grounds free-form commands into object-centric 3D state, interaction-relevant affordances, and collision-free motion with verification, reflection, and recovery.}
Experiments demonstrate consistent gains over both end-to-end VLA baselines and hierarchical open-vocabulary planners. UniManip achieves an average success rate of $0.94$ on the VLA benchmark (vs. $0.71$ for the strongest baseline) while substantially reducing distractor grasps, and reaches $0.83$ task success in cluttered scenes compared with ReKep, MOKA, and VoxPoser. On long-horizon and interaction-rich tasks, the full system attains $0.80$ average success over 90 trials in our ablation suite, highlighting the complementary roles of collision-free planning, relaxed IK, and AOG-driven recovery.
\textcolor{lblue}{The new graph-based comparison, no-graph control, LIBERO-Pro benchmark, and cross-embodiment trials sharpen the empirical scope of the contribution. Remaining failures are dominated by physical interaction uncertainty and embodiment-dependent kinematic constraints. Future work will incorporate tactile feedback and uncertainty-aware reconstruction, embodiment-aware whole-body planning, force/compliance control for contact-rich skills, and multi-arm coordination.}

\bibliographystyle{ieeetr}
\bibliography{refs}

\end{document}